\documentclass{article}

\usepackage{microtype}
\usepackage{graphicx}
\usepackage{booktabs} % for professional tables

\usepackage{hyperref}
\usepackage{amsmath}
\usepackage{amsfonts}
\usepackage[mathscr]{eucal}

\usepackage{afterpage}

\usepackage{tikz}
\usetikzlibrary{trees}
\usepackage{tabularx}

\usepackage{subcaption}
\usepackage[font={small,it}]{caption}
\usepackage{float}

\usepackage{xcolor}

\usepackage{algorithm}
\usepackage[noend]{algpseudocode}
\algdef{SE}[SUBALG]{Indent}{EndIndent}{}{\algorithmicend\ }%
\algtext*{Indent}
\algtext*{EndIndent}

\usepackage{latexsym}
\usepackage{caption}
\usepackage{listings}
\usepackage{float}
\usepackage{wrapfig}
\usepackage{multirow}
\usepackage{makecell}

\newcommand{\qed}{\nobreak \ifvmode \relax \else
    \ifdim\lastskip<1.5em \hskip-\lastskip
        \hskip1.5em plus0em minus0.5em \fi \nobreak
        \vrule height0.75em width0.5em depth0.25em\fi}

\usepackage{tikz}
\usepackage{verbatim}
\usepackage{scalefnt}
\usepackage{adjustbox}
\usepackage{tikz}
\usepackage{pgfplots}

\usetikzlibrary{calc,trees,positioning,arrows,chains,shapes.geometric,%
    decorations.pathreplacing,decorations.pathmorphing,shapes,%
    matrix,shapes.symbols,shadings}

\def\addlegendimage{\csname pgfplots@addlegendimage\endcsname}

\pgfkeys{/pgfplots/number in legend/.style={%
        /pgfplots/legend image code/.code={%
            \node at (0.125,-0.0225){#1}; % <= changed x value
        },%
    },
}
\pgfplotsset{
every legend to name picture/.style={west}
}

\definecolor{butter}{rgb}{0.98824, 0.91373, 0.30980}        % #fce94f
\definecolor{orange}{rgb}{0.98824, 0.68627, 0.24314}        % #fcaf3e

\definecolor{didi-gray}{cmyk}{0, 0, 0, .80}
\definecolor{didi-orange}{HTML}{FF7D41}

\definecolor{llgray}{RGB}{224, 224, 235}

\definecolor{didi-mint}{HTML}{3CDCBE}
\definecolor{didi-black}{cmyk}{0.5, 0.3, 0, .90}

\definecolor{didi-yellow}{HTML}{F9FC46}
\definecolor{didi-green}{HTML}{A7F26A}
\definecolor{didi-blue}{HTML}{69D2F1}

\colorlet{didi-l-blue}{white!80!didi-blue}
\colorlet{didi-ll-blue}{white!60!didi-blue}
\colorlet{didi-lll-blue}{white!40!didi-blue}
\colorlet{didi-llll-blue}{white!20!didi-blue}

\colorlet{didi-l-green}{white!80!didi-green}
\colorlet{didi-ll-green}{white!60!didi-green}
\colorlet{didi-lll-green}{white!40!didi-green}
\colorlet{didi-llll-green}{white!20!didi-green}

\colorlet{didi-l-yellow}{white!80!didi-yellow}
\colorlet{didi-ll-yellow}{white!60!didi-yellow}
\colorlet{didi-lll-yellow}{white!40!didi-yellow}
\colorlet{didi-llll-yellow}{white!20!didi-yellow}

\colorlet{didi-l-orange}{white!80!didi-orange}
\colorlet{didi-ll-orange}{white!60!didi-orange}
\colorlet{didi-lll-orange}{white!40!didi-orange}
\colorlet{didi-llll-orange}{white!20!didi-orange}

\colorlet{didi-l-gray}{white!20!didi-gray}
\colorlet{didi-ll-gray}{white!40!didi-gray}
\colorlet{didi-lll-gray}{white!60!didi-gray}
\colorlet{didi-llll-gray}{white!80!didi-gray}
\colorlet{didi-lllll-gray}{white!90!didi-gray}
\colorlet{didi-llllll-gray}{white!95!didi-gray}

\colorlet{didi-l-mint}{white!80!didi-mint}
\colorlet{didi-ll-mint}{white!60!didi-mint}
\colorlet{didi-lll-mint}{white!40!didi-mint}
\colorlet{didi-llll-mint}{white!20!didi-mint}

\usetikzlibrary{calc,trees,positioning,arrows,chains,shapes.geometric,%
    decorations.pathreplacing,decorations.pathmorphing,shapes,%
    matrix,shapes.symbols}

\usetikzlibrary{arrows.meta}

\tikzset{
    >={Latex[width=4mm,length=4mm]},
    base/.style = {rectangle, rounded corners, draw=black,
        text centered},
    portrait/.style = {minimum width=6mm, minimum height=4mm},
    transpose/.style = {minimum width=1cm, minimum height=3cm},
    data/.style = {base, portrait, fill=didi-llll-gray},
    discriminator/.style = {data, fill=didi-blue},
    data_t/.style = {data, transpose},
    output/.style = {base, portrait, fill=didi-lll-mint},
    output_t/.style = {output, transpose},
    process/.style = {base, portrait, 
        minimum width=2.5cm, 
        fill=didi-lll-orange,
    },
    process_t/.style = {process, transpose},
    process_t_2/.style = {process, transpose, fill=didi-ll-yellow},
    process_t_3/.style = {process, transpose, fill=didi-l-blue},
    to/.style={-latex, line width = 1pt},
    line_base/.style={->, color=didi-l-gray},
    to_dash/.style={line_base, dash dot, line width = 2pt},
    to_r_8/.style={->, line width = 8pt, color=didi-orange},
    to_r_4/.style={->, line width = 4pt, color=didi-orange},
    to_r_2/.style={->, line width = 2pt, color=didi-orange},
    to_r_2_dash/.style={->, dash dot, line width = 2pt, color=didi-orange},
    node/.style = {circle, draw = black, fill = didi-ll-mint},
}

\usepackage[accepted]{icml2021}

\icmltitlerunning{Policy-Gradient V2V GAN for Few-Shot Learning}

\pgfplotsset{compat=1.18}

\begin{document}

\twocolumn[
\icmltitle{Video to Video Generative Adversarial Network for Few-shot Learning Based on Policy Gradient}



\icmlsetsymbol{equal}{*}

\begin{icmlauthorlist}
\icmlauthor{Yintai Ma}{nwu}
\icmlauthor{Diego Klabjan}{nwu}
\icmlauthor{Jean Utke}{allstate}
\end{icmlauthorlist}
\icmlaffiliation{nwu}{Department of Industrial Enginering and Management Science, Northwestern University, Evanston, Illinois, United States}
\icmlaffiliation{allstate}{Allstate Insurance Company, Northbrook, Illinois, United States}
\icmlcorrespondingauthor{Yintai Ma}{yintaima2022@u.northwestern.edu}
\icmlcorrespondingauthor{Diego Klabjan}{d-klabjan@northwestern.edu}
\icmlcorrespondingauthor{Jean Utke}{jutke@allstate.com}
\icmlkeywords{Machine Learning}
\vskip 0.3in
]

\printAffiliationsAndNotice

\begin{abstract}
The development of sophisticated models for video-to-video synthesis has been facilitated by recent advances in deep reinforcement learning and generative adversarial networks (GANs). In this paper, we propose RL-V2V-GAN, a new deep neural network approach based on reinforcement learning for unsupervised conditional video-to-video synthesis. While preserving the unique style of the source video domain, our approach aims to learn a mapping from a source video domain to a target video domain. We train the model using policy gradient and employ ConvLSTM layers to capture the spatial and temporal information by designing a fine-grained GAN architecture and incorporating spatio-temporal adversarial goals. The adversarial losses aid in content translation while preserving style. Unlike traditional video-to-video synthesis methods requiring paired inputs, our proposed approach is more general because it does not require paired inputs. Thus, when dealing with limited videos in the target domain, i.e., few-shot learning, it is particularly effective. Our experiments show that RL-V2V-GAN can produce temporally coherent video results. These results highlight the potential of our approach for further advances in video-to-video synthesis.

\end{abstract}

\section{Introduction} \label{sch:intro} 

Video-to-video synthesis is a type of problem where a video is translated into another video while keeping some specific semantic meaning from the first video. Learning to synthesize continuous visual experiences is essential to build intelligent agents. For example, turning a daytime first-person car driving video into a nighttime driving video generates valuable samples for training autonomous driving agents. With such a learned video synthesis model, it would be possible to generate realistic videos without explicitly specifying scene geometry, materials, lighting, and dynamics, which would be cumbersome but necessary when using standard graphics rendering techniques. Learning to synthesize continuous visual videos serves both scientific interests and a wide range of applications including human motion and face translation from one person to another, teaching robots from human demonstration, or converting black-and-white videos to color. The ability to learn and model the temporal dynamics of our visual experience is also essential to building intelligent agents \citep{bandi_power_2023}.

While the image-based counterpart, the image-to-image synthesis problem, is a well-researched area, the video-to-video synthesis problem is less studied in existing works \citep{Zhuo2022}. The traditional frame-based models \citep{Cao2014, Thies2016}, in which the models take separate frames as inputs and outputs, ignore the temporal coherence within the videos and result in low visual quality and poor continuity. Most importantly, a frame-based model can not generate a correct sample based on its context or previous frames. Here we show an example in Figure \ref{fig:v2v_bad_sample}, where the goal is to translate an all-black background video to another video with the blue background in its first half and the red background in its second half. The target video has been obtained by the frame-based approach - and it clearly fails. Our proposed model outperforms the frame-based model because our video-to-video synthesis model correctly generates the background color for each output frame while the frame-based model randomly generates the background color for output frames. A more detailed and complex experiment is presented in Section \ref{sec:v2v:experiments}.

\newcolumntype{M}[1]{>{\centering\arraybackslash}m{#1}}

 \begin{figure}[htbp]
 \centering
 \begin{tabular}{M{2cm}M{5.5cm}}
\toprule
\textbf{Sequence Description } & \textbf{Frames in the Sequence }\\ 
\midrule
{Source video sequence}   &  \includegraphics[width=\linewidth]{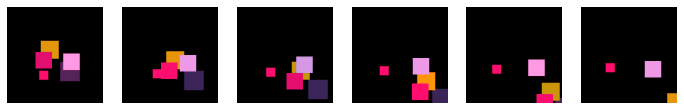}  \\
{Target video sequence} \hfil &  \includegraphics[width=\linewidth]{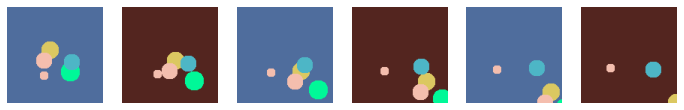}\\
\bottomrule
 \end{tabular} 
 \caption{The source video sequence is depicted in the first row and serves as the input to the model. The target video, shown in the second row, is characterized by a blue background in its first half and a red background in its second half.}
 \label{fig:v2v_bad_sample}
 \end{figure}

The image-to-image translation problem has been a popular topic in recent years and it has been widely studied
\citep{Isola2017, Taigman2017a, Bousmalis2017, Shrivastava2017, Zhu2017a, Liu2017,Liu2016,Huang2018,Zhu}. However, using these frame-based image-to-image models to translate videos has its inherent problem. As such models take separate images as inputs and outputs, they ignore the temporal coherence within the video and result in poor continuity and low visual quality. Only the models that include a recurrent neural network module or 3DCNN can better handle the video generation. Our proposed models directly model the temporal dynamics within videos with recurrent convolutional layers. We apply ConvLSTM to capture such temporal dynamics, which shows superior performance compared to the traditional way where the LSTM layer is attached after a CNN layer.

In this paper, we propose a reinforcement learning-based deep neural network approach, RL-V2V-GAN, for an unsupervised conditional video-to-video synthesis problem, for which the goal is to learn a mapping from a source video domain to another target video domain while preserving the style native to the target domain. The inputs are videos in the source domain, videos in the target domain, and videos and images in the target domain to serve as style references. The videos in the source domain have the same style as the style videos in the target domain. The goal is to create videos in the target domain that embody the common style by learning from the videos in the source domain.

There are numerous potential applications of RL-V2V-GAN. In e-commerce, it can transform generic product videos (source domain) into personalized recommendation videos (target domain). The common style is how the product is presented. This use case enhances customer engagement and conversion rates for businesses. In urban planning, it can transform abundant daytime cityscape videos (source domain) into mesmerizing nighttime aerial visuals (target domain). The common style is the appearance of buildings and the layout of cities. This aids city planners and promotes tourism by requiring less frequent data collection.

Our proposed video-to-video synthesis approach uses the generative adversarial neural network (GAN) framework. We design a custom GAN architecture, incorporating spatiotemporal adversarial objectives, and apply ConvLSTM layers to capture spatial and temporal information. Adversarial losses are employed to aid in the translation of content while preserving style. 

Within our proposed model, RL-V2V-GAN, we leverage the SeqGAN framework \cite{Yu2016}, by integrating Generative Adversarial Networks (GANs) with Reinforcement Learning (RL). Rather than utilizing traditional sequence generation methods, we employ RL, wherein the video generator functions as a stochastic policy. This approach facilitates a nuanced, contextual evaluation of each video frame within its encompassing sequence, ensuring superior temporal coherence. Derived from the GAN discriminator, the reward signals provide crucial feedback for optimization of synthetic video generation. This reliance on RL is pivotal for our model's capability to generate video sequences that closely align with the stylistic attributes of the target domain.

This approach is completely unsupervised since it does not require paired inputs, making it particularly effective for zero or few-shot learning. The experiments demonstrate that our model is capable of generating temporally coherent video results based solely on input visual frames. By leveraging the proven generative adversarial framework of CycleGAN \citep{Wexler2007}, our innovative approach offers unsupervised training of video-to-video synthesis models, enabling the translation of videos to another domain with one domain of training samples
and a few in the target domain, resulting in a wide range of visually appealing output videos.
 
The main contribution of this paper is the first deep reinforcement learning algorithm with a recurrent neural network approach that solves a conditional video-to-video synthesis problem. This novel approach seamlessly integrates reinforcement learning (RL) with generative adversarial networks (GANs), leveraging the strengths of both to achieve temporal coherence and stylistic fidelity in generated video sequences. The existing deep neural network models for conditional video synthesis tasks are frame-based, where each frame in the translated video is generated separately. These types of models suffer when generating a high-quality video over a long period. However, our model natively takes consecutive frames as one single input. It helps to generate videos where the content in each frame is temporally coherent. 
    
This paper is organized as follows. In Section \ref{sec:v2v:lr}, we review the related works and the development of the video retrieval methods. We formally state our model and algorithm in Section \ref{sec:v2v:model}. We discuss our algorithm in Section \ref{sec:alg}. We present our numerical experiments results in Sections \ref{sec:v2v:experiments}. 

\section{Related Works}
\label{sec:v2v:lr}
In this section, we discuss advancements in video-to-video synthesis.

\textbf{Video Synthesis}
Video synthesis can be broadly categorized into unconditional and conditional approaches, alongside a refined classification based on temporal dimension handling, namely: frame-to-frame, frame-to-video, and video-to-video synthesis.

Unconditional video synthesis operates without explicit input beyond random noise, aiming to generate coherent video content. Pioneering works \citep{Carl2016, Saito2017, Tulyakov2018} extend the traditional GAN framework for unconditional video synthesis, focusing on the conversion of random vectors into video sequences through spatio-temporal convolutional networks and latent image code projection. \citet{Saito2017} propose TGAN to project a latent vector to a set of latent image codes, and then convert these latent image codes to frames with an image generator.

Conversely, conditional video synthesis generates videos based on given inputs or conditions, such as semantic representations or single frames. This approach often leverages temporal models for predicting future poses or generating videos conditioned on specific inputs, as seen in works such as \citet{He2018}, \citet{Villegas2017}, \citet{Walker2017}, and further advancements by \citet{Mathieu2016} and \citet{chen_rethinking_2017}. Our method falls into this category, employing ConvLSTM layers within a GAN framework to capture spatial and temporal dynamics, ensuring high fidelity in video synthesis.

Diving deeper into the temporal handling categorization, the frame-to-frame video synthesis models take each frame as input for its generator and are mostly CNN-based. They usually include a discriminator in the model to decide whether the generated sequence is a video \citep{Chen2017,Gupta2017,Huang2017,Ruder2016}. 

The second type of frame-to-video synthesis models try to generate a sequence of frames based on one single input frame. \citet{Tulyakov2018}'s MoCoGAN separates the latent space to motion and content subspaces and uses an RNN generator to obtain a sequence of motion embeddings. However, it often struggles with high-resolution or long videos due to the complexity of modeling motion dynamics.

The last video-to-video models are the most advanced ones that directly take temporal patterns as part of the inputs. They usually take the entire video 3-dimension tensor as input. \citet{Bansal2018} propose a video-to-video translation model with ReCycleGAN that includes a cross-domain cyclic loss for temporally coherent frames. \citet{Wang2018} propose a conditional video-to-video synthesis model with a sequential image-based generator. \citet{shen_mostgan-v_2023} present MoStGAN-V, which uses temporal motion styles to model diverse and temporally-consistent motions in video generation. This technique enhances the generation of dynamic motions, relevant to our model's objectives. \citet{ma_cvegan_2024} introduce CVEGAN, a GAN-based model designed for enhancing the quality of compressed video frames. This approach is relevant for applications requiring high-quality video outputs, aligning with our objectives of generating visually coherent videos.

Generally speaking, the videos generated by the above image-to-image models suffer from temporal incoherence. In contrast to the first two categories, we focus on designing a video-to-video translation approach that generates videos with coherent frames. All the above models process videos frame by frame while we generate an entire video in one shot. Unlike traditional methods that heavily rely on paired inputs, our model's unsupervised nature makes it exceptionally resilient and adaptable, proving invaluable especially when faced with limited samples in the target domain.

\textbf{SeqGAN} \citep{Yu2016} is a sequence generation framework that leverages Generative Adversarial Networks (GAN) to train a generative model. Traditional GANs have limitations when generating sequences of discrete tokens due to the difficulties in passing gradient updates from the discriminative model to the generative model. To overcome this, SeqGAN models the data generator as a stochastic policy in reinforcement learning and performs gradient policy updates directly, bypassing the generator differentiation problem. The RL reward signal is obtained from the GAN discriminator, which is trained to provide positive examples from real sequence data and negative examples from synthetic sequences generated from the generative model. The reward signal is passed back to the intermediate state-action steps using a Monte Carlo search. The policy gradient is calculated based on the expected end reward received from the discriminative model, which represents the likelihood of fooling the discriminator. Extensive experiments on synthetic data and real-world tasks demonstrate the effectiveness of SeqGAN. Based on the advancements of SeqGAN, our work generates the entire video sequence with Policy Gradient. RL-V2V-GAN represents an innovative convergence of deep reinforcement learning with GAN for video synthesis. This strategy not only yields temporally coherent outputs but also comprehends and replicates the intricate stylistic elements inherent to the source domain.

\textbf{Generative AI and Diffusion Models} Generative AI, particularly through advancements in diffusion models, has revolutionized the field of artificial intelligence by enabling the generation of high-quality, realistic data across various domains. 
\citet{yu_video_2023} propose Projected Latent Video Diffusion Models (PVDM), which efficiently handle high-resolution video synthesis in a low-dimensional latent space. This method addresses the challenges of high-dimensionality and complex temporal dynamics, which is pertinent to our work.
\citet{yang_diffusion_2023} present a diffusion probabilistic model for video generation that improves perceptual quality and probabilistic frame forecasting. Their approach is relevant for ensuring temporal coherence in generated videos, similar to our objectives.
\citet{chen_seine_2023} introduce SEINE, a short-to-long video diffusion model focusing on generative transition and prediction, which can be extended to various tasks such as image-to-video animation. This approach is relevant to enhancing the temporal dynamics in video generation.
\citet{esser_structure_2023} present a structure and content-guided video diffusion model that edits videos based on user descriptions, providing fine-grained control over output characteristics and ensuring temporal consistency. This work aligns with our goal of generating high-fidelity, temporally coherent videos.
\citet{kim_diffusion_2023} propose a face video editing framework based on diffusion autoencoders that ensures temporal consistency. This method is pertinent to our goal of maintaining coherence in video synthesis.
\citet{zeng_make_2024} present PixelDance, a novel approach for high-dynamic video generation using diffusion models. This method's success in synthesizing complex scenes and motions is relevant to enhancing the visual dynamics in our video generation.

These approaches primarily utilize diffusion models, whereas our work focuses on GAN-based video synthesis. Generative Adversarial Networks (GANs) and diffusion models each have distinct advantages. GANs excel in generating high-quality, realistic images quickly once trained but can suffer from training instability and mode collapse. Diffusion models offer stable training and diverse outputs but are slower during inference due to their iterative process. GANs are ideal for tasks demanding efficient, high-quality image generation, while diffusion models suit applications requiring stability and diversity. Our work focuses on GAN-based video synthesis and uniquely integrates reinforcement learning to enhance training efficiency and video quality. By leveraging GANs' strengths, our approach ensures that the generated videos are both realistic and stylistically consistent with the target domain.

\textbf{Video Generation with Reinforcement Learning}
has seen significant advancements in recent years. Notably, the VIPER framework \citep{escontrela_video_2023} leverages video prediction models to generate reward signals for training RL agents. Both VIPER and our approach share common ground in utilizing generative models for video and integrating RL to enhance the learning process. However, VIPER focuses on policy learning by using video model log-likelihoods as reward signals, whereas our RL-V2V-GAN directly addresses the challenge of video-to-video synthesis, generating high-quality, temporally coherent videos. Our unique contribution lies in the novel integration of RL with GAN for unsupervised video synthesis. Unlike VIPER, which aims to train RL agents by mimicking expert trajectories, our approach employs spatio-temporal adversarial objectives and ConvLSTM layers to capture both spatial and temporal information, ensuring the stylistic fidelity and coherence of generated videos. Additionally, our model's ability to perform few-shot learning makes it particularly effective for generating videos in scenarios with limited target domain data. The specific settings that enable our model to solve the problem effectively include using ConvLSTM layers to capture temporal dynamics, applying spatio-temporal adversarial objectives, and employing a robust combination of adversarial, recurrent, recycle, and video losses.

\section{Model}
\label{sec:v2v:model}

This section outlines the mathematical notation and models to accurately specify the elements of RL-V2V-GAN. We start by establishing notation and proceed to explain the model's GAN architecture, including its generators, discriminators, and predictors. We emphasize the model's loss functions. The role of RL in refining video synthesis is then highlighted. Finally, we outline the key neural network components, such as ConvLSTM layers and neural blocks, essential for video generation. This framework facilitates the translation of videos across various domains and styles.

\subsection{Notation}

We define $\mathcal X$ and $\mathcal Y$ to be sets of videos $\boldsymbol x\in \mathcal X$ and $\boldsymbol y \in \mathcal Y$ in the source and target domain, respectively. The notation $T$ denotes the length of a video, and subscripting $\boldsymbol x_i$ selects the $i$-th frame from video $\boldsymbol x$.
Notations $\boldsymbol x_{:j}$ and $\boldsymbol y_{:j}$ represent videos from frames $1$ to $j$. We also use $x$ and $y$ to represent a single frame from the source or target domain.  Similarly, we define $\boldsymbol z\in \mathcal Z$ to represent a style video in the target domain, and $\bar{z}\in \bar{\mathcal Z}$ to represent a style image in the target domain. Styles represented by $\mathcal Z$ and $\bar{\mathcal Z}$ are assumed to be compatible. 

\subsection{Generative Adversarial Network Model}
In this work, the proposed model creates additional videos $\boldsymbol{y} \in \mathcal Y$ that incorporate features from both source videos $\boldsymbol{x}\in \mathcal X$ and style of $\boldsymbol{z}\in \mathcal Z$.  Our goal is to overcome the challenge of having a much smaller number of videos $\boldsymbol{y}$ for the desired style and domain. The model takes input videos from $\mathcal X, \mathcal Y, \mathcal Z$, as well as images of desired style $\bar{z}\in\bar{\mathcal Z}$. With these inputs, our model generates new videos in the style of $\mathcal Z$ and $\bar{\mathcal Z}$, addressing the scarcity of videos in $\mathcal Y$  in the preferred style.

This GAN model contains sequence generators $G_{x}$, $G_{y}$, predictors $P_{x}$, $P_{y}$ and discriminators $D_{x}$, $D_{y}$. Here $x$ is always from the source domain, $y$ is always from the target domain and $z$ is always a style frame from the target domain. The generators map a video from the source domain to a video in the target domain or from the target domain to the source domain, respectively. 

We define $G_y: $  $\boldsymbol{x} \rightarrow \boldsymbol{y}$ and $G_x: $ $\boldsymbol{y} \rightarrow \boldsymbol{x}$. The predictor uses a video to predict the next frame of this video in the same domain, for the source and target domain, respectively. We define $P_x: $ $ \boldsymbol{x}_{:t} \rightarrow \boldsymbol{x}_{t+1} $ and $P_y: \boldsymbol{y}_{:t} \rightarrow \boldsymbol{y}_{t+1} $. Although the predictor $P$ has the same sequence-to-sequence neural network architecture as the video generator $G$, we denote $P_x(\cdot)$ and $P_y(\cdot)$ to represent the prediction of the next frame, rather than the entire output sequence. In our model, the generator produces three logits per pixel, corresponding to the RGB channels. These logits are not independent as they are generated using ConvLSTM layers, which capture spatial and temporal dependencies. This approach is similar to an LSTM but operates on the convolutional feature maps, ensuring that the temporal coherence between video frames is maintained. The sequence discriminators have two use cases; they can take either a frame or a sequence of frames as input. When a discriminator takes one frame as input, we define $D_x: x \rightarrow [0,1] $ and $D_y: $ $ y \rightarrow [0,1]^2 $. In this case, $D_y$ produces two outputs: $D_{y,0}$ predicts whether the sample sequence belongs to the target domain, and $D_{y,1}$ predicts whether the sample has the desired style, based on adversarial learning with samples from \( \mathcal{Z} \) or \( \bar{\mathcal{Z}} \). Positive samples for $D_{y,1}$ include video data from $z \in \mathcal{Z}$, while generated samples are negative. In the former case, when a video is fed to a source domain discriminator, we define $D_{\boldsymbol{x}}: \boldsymbol{x} \rightarrow [0,1] $ that outputs a probability indicating whether the video sequence belongs to the source domain. When a video is fed to a target domain discriminator, $D_{\boldsymbol{y}}$ outputs two scalar values, denoted as $[D_{\boldsymbol{y},0}, D_{\boldsymbol{y},1}] \in [0,1]^2$. The probability $D_{\boldsymbol{y},0}$ specifies whether the sample sequence belongs to the target domain, and $D_{\boldsymbol{y},1}$ indicates whether the sample sequence has the desired style. For an example of how the datasets \( \mathcal{X} \), \( \mathcal{Y} \), and \( \mathcal{Z} \) are structured, please refer to Table \ref{table-datasets-city}, which illustrates data from modern cities in daytime and small-town videos in both daytime and nighttime, along with style references from small-town daytime footage and modern city nighttime images.

Next we list loss components.

\subsubsection{Adversarial loss $L_g$}

With video samples $\{\boldsymbol{x}_{:t}\} \in \mathcal{X}$, $\{\boldsymbol{y}_{:t}\} \in \mathcal{Y}$, and $\{\boldsymbol{z}_{:t}\} \in \mathcal{Z}$, as well as frame samples $\{\bar{y}\} \in \bar{\mathcal{Z}}$, $\boldsymbol{y}_t \in \{\boldsymbol{y}_{:t}\}$ and $\boldsymbol{z}_t \in \{\boldsymbol{z}_{:t}\}$, we define the standard adversarial loss to distinguish between true samples and synthetic samples generated by the model. Here, $\bar{y}$ represents an image with the desired style. 

The objective functions are defined using the adversarial losses $L_g^y$ and $L_g^x$, where the generator minimizes these losses and the discriminator maximizes them, following the standard minimax framework of GANs. Although the log terms in the loss are negative, minimizing the generator’s loss is equivalent to maximizing the likelihood of generating realistic data. This setup aligns with common GAN formulations, where both the generator and discriminator operate within the output range of $[0,1]$.

\begin{equation}
\begin{array}{rl}
&\min_{G_y} \max_{D_y} L^y_g(G_y, D_y) :=  \\
& \sum_{s} \log D_{y,0}(\bar{y}_s) +  \sum_{t} \log D_{y,1}(\boldsymbol{z}_t) \\
\phantom{=}& + \sum_{t} \log D_{y,0}(\boldsymbol{y}_t) + \sum_{t} \log (1-D_{y,0}((G_y(\boldsymbol{x}_{:t}))_{t}))
\end{array}
\end{equation}

\begin{equation}
\begin{array}{rl}
&\min_{G_x} \max_{D_x} L^x_g(G_x, D_x) := \\
& \sum_{t} \log D_x(\boldsymbol{x}_t) + \sum_{t} \log (1-D_{x}((G_x(\boldsymbol{y}_{:t}))_{t})) \\
\phantom{=}& + \sum_{t} \log (1-D_{x}((G_x(\boldsymbol{z}_{:t}))_{t}))
\end{array}
\end{equation}

\subsubsection{Recurrent Loss $L_{rr}$}

By encouraging the model to generate more temporally coherent frames by a recurrent loss, we can impose more advantage of the temporal ordering. We embed the recurrent temporal predictor $P_x$ and $P_y$ with the recurrent loss $L^x_{rr}$ and $L^y_{rr}$, respectively.

\begin{equation}
    L^x_{rr} (P_x) = \sum_t \| \boldsymbol{x}_{t+1} - P_x(\boldsymbol{x}_{:t}) \|^2
\end{equation}

\begin{equation}
    L^y_{rr} (P_y) = \sum_t \| \boldsymbol{y}_{t+1} - P_y(\boldsymbol{y}_{:t}) \|^2 + \sum_t \| \boldsymbol{z}_{t+1} - P_y(\boldsymbol{z}_{:t}) \|^2
\end{equation}

\subsubsection{ReCycle Loss $L_{rc}$}

We introduce the recycle loss $L_{rc}$, which encourages style and domain consistency when multiple generators and predictors are jointly utilized, crossing domains and time altogether. For example, we start with a video $\boldsymbol{x}_{:t}$, translate it to a video in the target domain denoted by $G_y(\boldsymbol{x}_{:t})$, then increment it on time dimension as $P_y(  G_y(\boldsymbol{x}_{:t}) ))$. We select the last frame, which is the $t+1$-st frame in the source domain with subscript $t+1$ as $G_x((\boldsymbol{y}_{:t},P_y(G_y(\boldsymbol{x}_{:t})) ))_{t+1}$ and compare this frame with the ground truth $x_{t+1}$ in a L-2 loss.

\begin{equation}
\begin{array}{rl}
    &L^{xy}_{rc}(G_x, G_y, P_y)  \\
    =& \sum_t \| \boldsymbol{x}_{t+1} - G_x((\boldsymbol{y}_{:t},P_y(G_y(\boldsymbol{x}_{:t}))))_{t+1} \|^2
\end{array}
\end{equation}
and 
\begin{equation}
\begin{array}{rl}
    & L^{yx}_{rc}(G_y, G_x, P_x)  \\
    =&  \sum_t \lVert \boldsymbol{y}_{t+1} - G_y((\boldsymbol{x}_{:t},P_x(G_x(\boldsymbol{y}_{:t}))))_{t+1} \rVert^2 + \\
     & \sum_t \lVert \boldsymbol{z}_{t+1} - G_y((\boldsymbol{x}_{:t},P_x(G_x(\boldsymbol{z}_{:t}))))_{t+1} \rVert^2
\end{array}
\end{equation}

\subsubsection{Video Loss $L_v$}
Finally, we leverage the specially designed discriminator capable of processing both individual frames and full videos. We design video loss $L_v$ to supervise the overall style and quality of the generated video for the target domain. The purpose is to ensure the consecutive output frames resemble the temporal dynamics of a real video. The video loss $L_v$ is given by:

\begin{equation}
\begin{array}{rl}
      & L_v(G_x, G_y, D_x, D_y) \\
    = & L_{vx}(G_x, G_y, D_{x}) + L_{vy}(G_y, G_x, D_{y}),
\end{array}
\end{equation}
where
\begin{equation}
    \begin{array}{rl}
        & L_{vx}(G_x, G_y, D_{x}) \\
      = & \sum_t \log D_{\boldsymbol{x}}(\boldsymbol{x}_{:t}) +  \\ 
      + & \sum_t \log(1- D_{\boldsymbol{x}}(G_x(\boldsymbol{y}_{:t}))) \\
      + & \sum_t \log(1- D_{\boldsymbol{x}}(G_x(\boldsymbol{z}_{:t}))).
    \end{array}
\end{equation}
and 
\begin{equation}
    \begin{array}{rl}
         &  L_{vy}(G_y, G_x, D_{y}) \\
       = &  \sum_t \log D_{\boldsymbol{y},0}(\boldsymbol{y}_{:t}) \\
       + &  \sum_t \log D_{\boldsymbol{y},1}(\boldsymbol{z}_{:t}) \\
       + &  \sum_t \log(1- D_{\boldsymbol{y},0}(G_y(\boldsymbol{x}_{:t}))) \\
       + &  \sum_t \log(1- D_{\boldsymbol{y},1}(G_y(\boldsymbol{x}_{:t}))).
    \end{array}
\end{equation}

\subsubsection{Total loss of RL-V2V-GAN}

We now combine the above losses and rewrite it in a standard min-max formulation to form the total loss used for the proposed RL-V2V-GAN model:

% TODO explain what lambda is here

\begin{equation}
    \begin{array}{rl}
      & \min_{G,P} \max_{D} L(G,P,D) \\
    = & L^x_g(G_x, D_x) + L^y_g(G_y, D_y)  \\ 
      & + \lambda_{rrx} L^x_{rr} (P_x) + \lambda_{rry} L^y_{rr} (P_y) ) \\ 
      & + \lambda_{rcx} L^{xy}_{rc} (G_x, G_y, P_y) + \lambda_{rcy} L^{yx}_{rc} (G_y, G_x, P_x)    )  \\ 
      & + \lambda_{vx} L_{vx}(G_x, G_y, D_{x}) )  + \lambda_{vy} L_{vy}(G_x, G_y, D_{y}) ) 
    \end{array}
\end{equation}
The gradients of this loss can be separated into two parts. 
% The first part, $\partial L / \partial G+ \partial L / \partial P$, updates the generator and predictor with a deterministic policy gradient. 
The first part, $\partial (L-L_v) / \partial D$, trains discriminators for non-terminating states. The second part, $\partial L_v / \partial D$, trains discriminators for video input using terminating state rewards. Gradient penalty coefficients $\lambda=\{\lambda_{vx}, \lambda_{vy}, \lambda_{rrx}, \lambda_{rry}, \lambda_{rcx}, \lambda_{rcy}\}$ are added to control the relative importance of each loss components during training.

\begin{figure}[H]
    \centering
    \begin{subfigure}[b]{.5\textwidth}
    \centering
    \includegraphics[width=\linewidth]{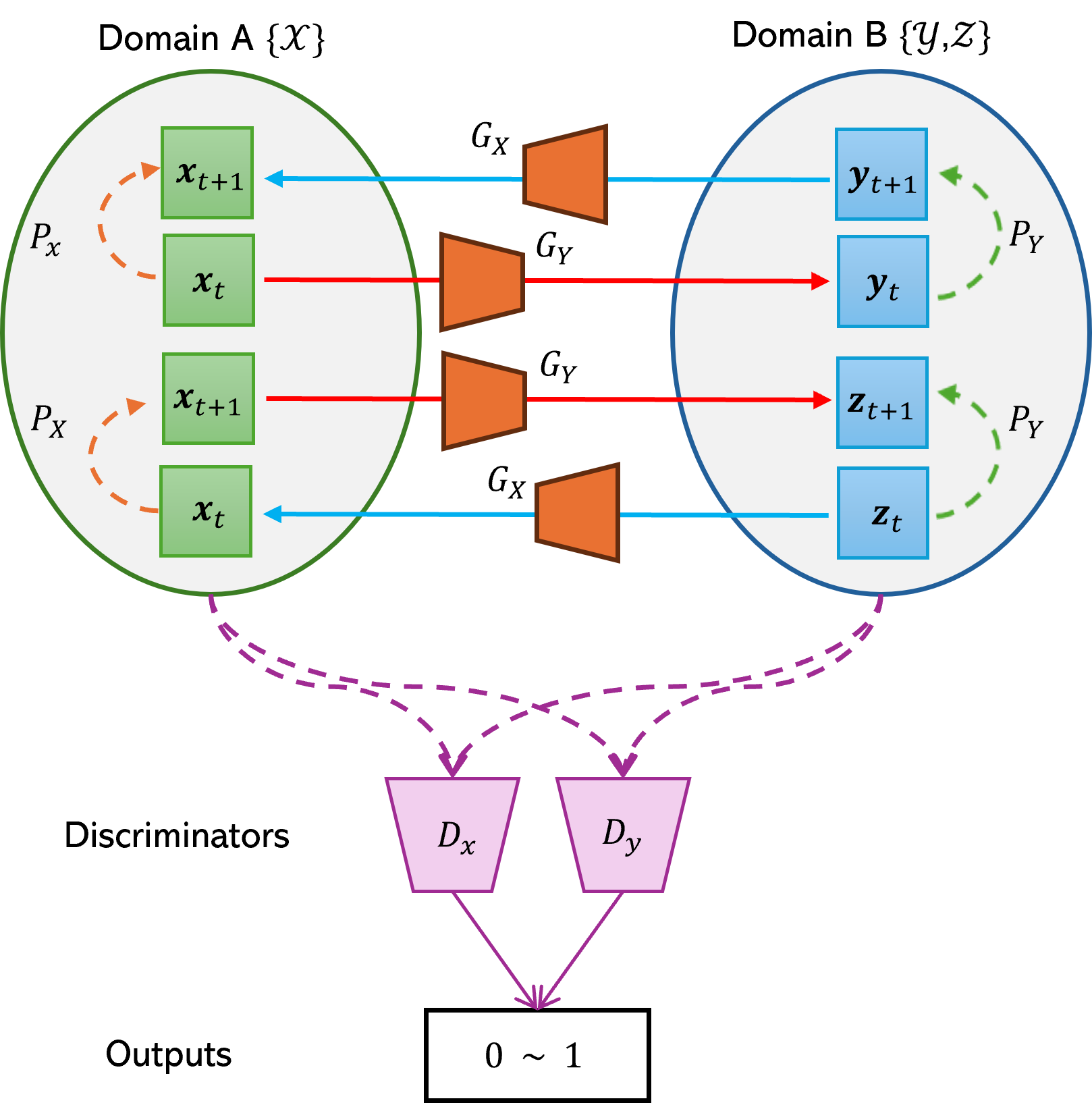}
    \end{subfigure}
    \caption{ The diagram presents the RL-V2V-GAN model, which integrates sequence generators $G_x$, $G_y$, predictors $P_x$, $P_y$, and discriminators $D_x$. $D_y$ for video style transfer. It captures the workflow where $G$ networks transform videos between domains, $P$ networks forecast future frames and $D$ networks assess authenticity and style. The model operates under various losses—adversarial, recurrent, ReCycle, and video—to ensure high-quality, coherent video generation, addressing the challenge of data scarcity in style-specific videos.}
    \label{fig:model_GAN}
\end{figure}

\subsection{Reinforcement Learning Model}

For the reinforcement learning part of the algorithm, we define two Q-networks $Q=\{Q_x(s,a), Q_y(s,a)\}$ and the policy networks $\mu= \{G_x(s), G_y(s), P_x(s), P_y(s)\}$. The state $s$ is defined as a video and action $a$ is a frame that could happen right after such a video. Given a video and a future frame, the Q-networks generate a reward $\hat{r}$ to estimate how likely this frame could happen as the next frame to this video, i.e., $Q(s,a) \rightarrow \hat{r}$. We have one Q network for each domain, such that $Q_x(s,a)$ evaluates how good the action is for the source domain, and $Q_y(s,a)$ evaluates the action for the target domain. They are parameterized by $\theta^Q$. Each policy network in $\mu$ outputs the action (next frame) for a given state $s$ (video). All four of the generators $G_x$, $G_y$, predictors $P_x$ and $P_y$ are policy networks parameterized by $\theta^\mu = \{\theta^{G_{x}}, \theta^{G_{y}}, \theta^{P_{x}}, \theta^{P_{x}} \}$.

In the reinforcement learning training process, we characterize each transition by $(s, a, r, s', a^{\text{true}}, p)$, where $a^{\text{true}}$ represents the actual subsequent frame following the sequence $s$. The positional variable $p$ indicates whether the action frame has reached the maximum number of frames in a video, denoted as $T$. The length of each video is fixed at $T$ frames. The reward $r$ is computed based on whether the state has reached the end of a video. When this terminal state is reached (i.e., $p = T - 1$), the reward is obtained from the video loss: $r \leftarrow - \lambda_v L_v$. Otherwise, the reward is calculated from the adversarial, recurrent, and recycle losses: $r_t \leftarrow -(L_g + \lambda_{rr} L_{rr} + \lambda_{rc} L_{rc})$. This ensures that the reward captures both the quality of the generated frame and its temporal coherence with previous frames.

To enhance stability and facilitate effective learning, transitions are continuously collected in a replay buffer $\mathcal{B}$ during training. This buffer is crucial for implementing delayed policy updates. New transitions are added until the buffer reaches its capacity limit, at which point older experiences are replaced by newer ones. This ensures a diverse set of experiences for sampling mini-batches, which are then used to update the policy networks.

To improve training stability and reduce fluctuations, our model employs target networks, $Q'$ and $\mu'$, as stable versions of Q-networks and policy networks, respectively. These target networks update their weights gradually, mirroring the primary networks at specific intervals. This strategy, rooted in deep Q-learning techniques, ensures smoother training by offering a consistent benchmark for policy evaluation and reward estimation, thereby minimizing prediction variability and enhancing learning reliability. We also refer to the target networks with its parameter as $\theta^{Q'}$ and $\theta^{\mu'}$.

\begin{figure*}[t]
  \centering
  \includegraphics[width=\textwidth]{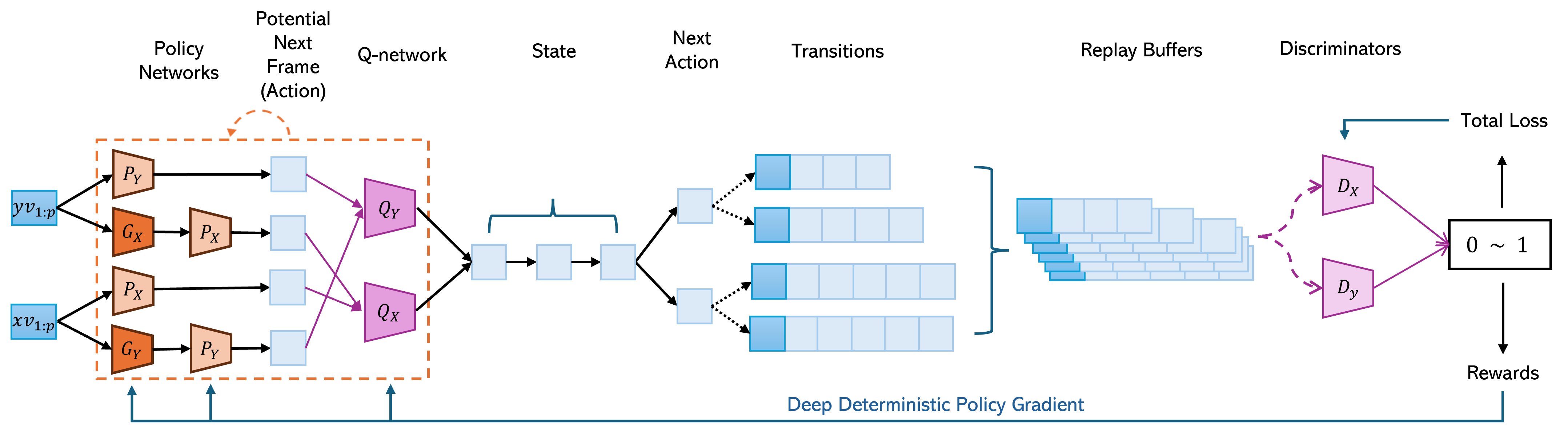}
  \caption{
  This figure showcases the reinforcement learning mechanism of RL-V2V-GAN, involving Q-networks ($Q_x$, $Q_y$) and policy networks ($\mu$). States ($s$) are videos, and actions ($a$) are potential frames. Q-networks assess the flow of frame sequences, guiding the model to produce coherent and stylistically accurate videos. The system collects transitions in a replay buffer, optimizing for actions that yield realistic sequences and updates policy and Q-network with policy gradient.}
  \label{fig:modeL_RL}
\end{figure*}

\subsection{Neural Network Components}

\begin{figure}[htbp]
    \centering
        \begin{subfigure}[b]{0.14\textwidth}
                \centering
                \includegraphics[width=1\linewidth]{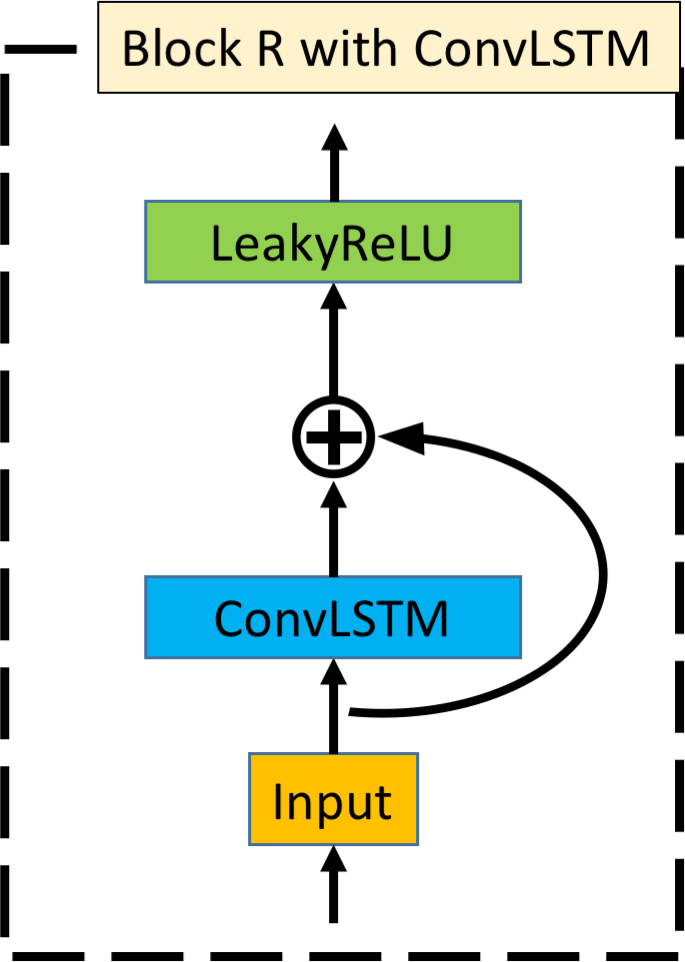}
                \caption{Block R}
        \label{fig:v2v_r}
        \end{subfigure}
    \begin{subfigure}[b]{0.14\textwidth}
            \centering
            \includegraphics[width=\linewidth]{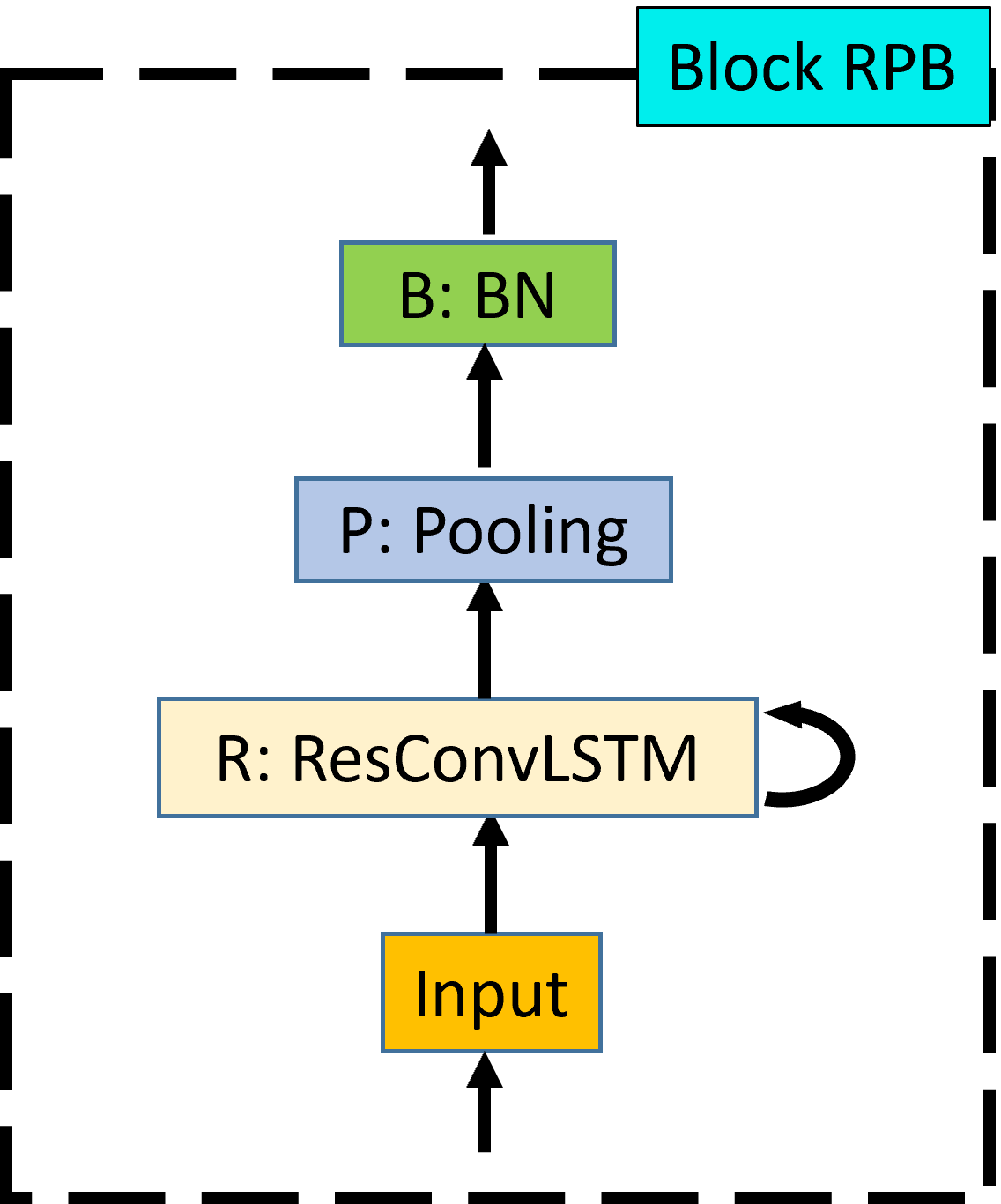}
            \caption{Block RPB}
    \label{fig:v2v_rbp}
    \end{subfigure}
    \begin{subfigure}[b]{0.14\textwidth}
            \centering 
            \includegraphics[width=\linewidth]{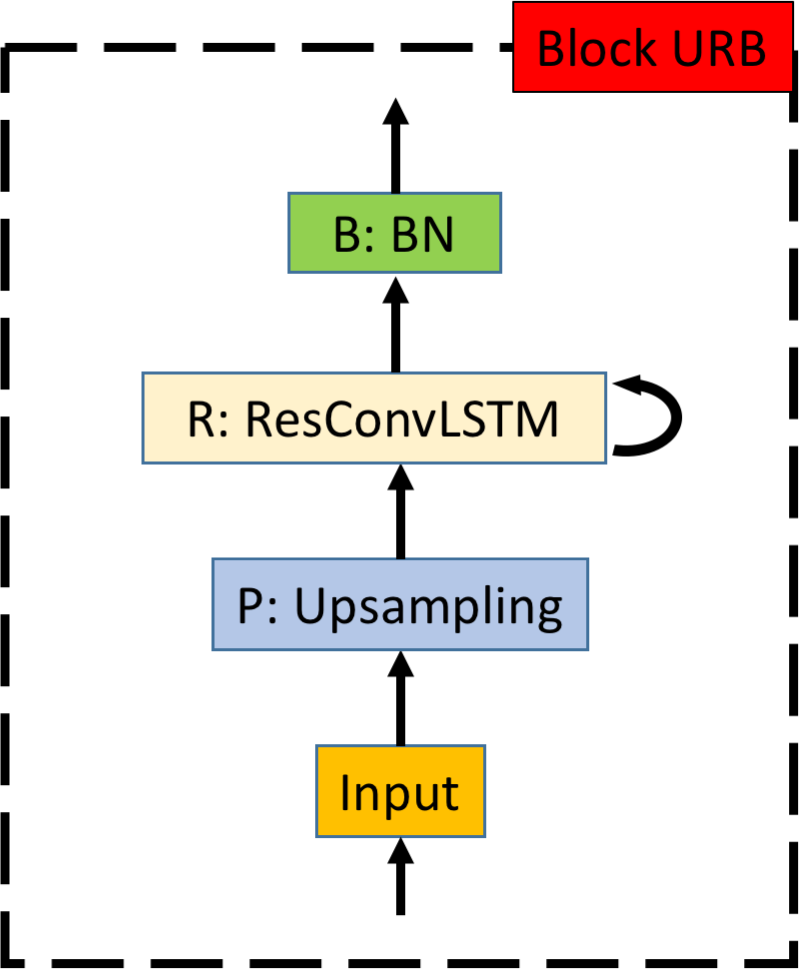}
            \caption{Block URB}
    \label{fig:v2v_block_urb}
    \end{subfigure}
        
    \caption{ 
    (\protect\subref{fig:v2v_r}) shows the structure of block R.
    (\protect\subref{fig:v2v_rbp}) shows the structure of block RPB. 
    (\protect\subref{fig:v2v_block_urb}) shows the structure of block URB.}
    \label{fig:v2v_basic_blocks}
\end{figure}

In our RL-V2V-GAN model, both the generators $G_x$, $G_y$ and predictors $P_x$ and $P_y$ are sequence to sequence auto-encoders. They share the same architecture, constructed by three different core neural network blocks: block R, RPB, and URB. Illustrated in Figures \ref{fig:v2v_r}, \ref{fig:v2v_rbp}, and \ref{fig:v2v_block_urb}, each block plays a pivotal role in processing the video sequences. The ConvLSTM layers, integrated within these blocks, are enhanced with residual connections, pooling layers, and batch normalization layers to capture both spatial and temporal dynamics effectively. Among these, the R block (Residual), featuring a LeakyReLU activation function, stands as the foundation, ensuring efficient data flow through the model. The RPB block (Residual-Pooling-BatchNormalization) serves a critical function in compressing the input dimensions, facilitating a more manageable representation of the data. Conversely, the URB block (Upsampling-Residual-BatchNormalization) is tasked with expanding these compressed inputs back to their higher dimensional form, thus preserving the integrity of the video's original structure and detail.

\begin{figure}[htbp]
    \centering
    \begin{subfigure}[b]{.13\textwidth}
            \centering
            \includegraphics[width=\linewidth]{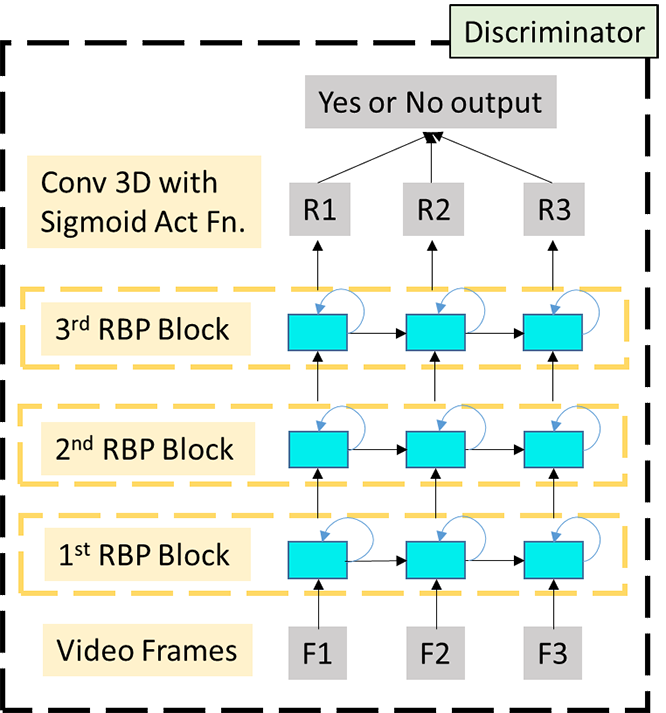}
            \caption{\small Discriminator}
            \label{fig:v2v_dis}
    \end{subfigure}
    \begin{subfigure}[b]{.32\textwidth}
    \centering
    \includegraphics[width=\linewidth]{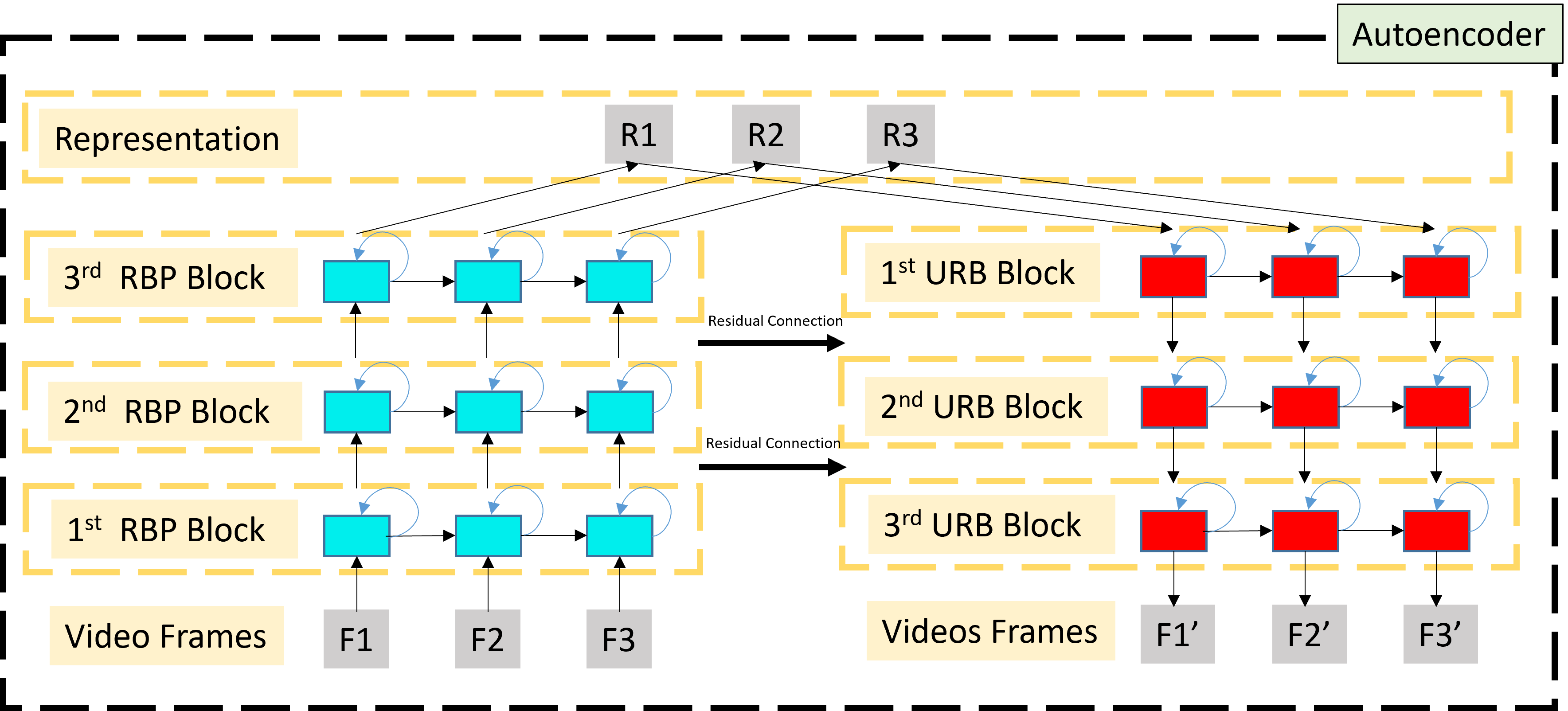}
    \caption{Generator}
    \label{fig:v2v_gen}
    \end{subfigure}
        
    \caption{ (\protect\subref{fig:v2v_dis}) shows the structure of a discriminator. This instance contains three layers of RPB blocks and a 3D convolutional layer with a sigmoid activation function. (\protect\subref{fig:v2v_gen}) shows the structure of a generator. This instance contains three layers of RPB blocks in the encoder and three layers of URB blocks in the decoder, respectively.}
    \label{fig:v2v_dis_and_gen}
\end{figure}

With the definitions of block R, RPB, and URB in place, we can now describe the architecture of our generators and discriminators in Figure \ref{fig:v2v_dis_and_gen}. Our discriminator contains a sequence of RPB blocks, followed by a 3D convolutional layer that maps the video input to a single binary decision. On the other hand, the generator consists of an encoding part composed of RPB blocks and a decoding part made up of URB blocks. However, it is worth noting that the embedding vector obtained from the encoding part is not a standalone representation of the video, as there are residual connections established between each corresponding pair of RPB-URB layers.

\section{Model Configuration}

The RL-V2V-GAN model employs a unified architecture across its network components for streamlined video synthesis. This design ensures that the generated videos maintain temporal coherence and high visual quality. The configuration of the model is detailed in Table \ref{table-model-architecture}. All four networks share the same encoder architecture of 3 RPB layers. Both generators $G$ and predictors $P$ use the same decoder structure of 3 URB block layers to generate video as output. Similarly, both discriminators $D$ and Q-networks $Q$ have the same decoder structure of 2 RPB block layers and 2 fully connected (FC) layers to generate scalar outputs for effective video content analysis and reward estimation. The Q-network uses ReLU in the final fully connected layer to allow for a wide range of output values, while the discriminator uses Sigmoid to output probabilities.

\begin{table}[H]
\caption{The encoder and decoder blocks in all networks}
\label{table-model-architecture}
% \vskip 0.05in
\begin{center}
\begin{scriptsize}
\begin{sc}
\begin{tabular}{lllc}
\toprule
Network & Encoder & Decoder & Final Func. \\
\midrule
Generators ($G$)     & 3 RPB  & 3 URB       & ConvLSTM \\
Predictors ($P$)     & 3 RPB  & 3 URB       & ConvLSTM \\
Discriminators ($D$) & 3 RPB  & 2 RPB-FC-FC & Sigmoid  \\
Q-networks ($Q$)     & 3 RPB  & 2 RPB-FC-FC & ReLU     \\
\bottomrule
\end{tabular}
\end{sc}
\end{scriptsize}
\end{center}
\vskip -0.1in
\end{table}

The RL-V2V-GAN model trains four policy networks ($G_x$, $G_y$, $P_x$, $P_y$), two discriminators ($D_x$, $D_y$), and two Q-networks ($Q_x$, $Q_y$). These networks are initialized with pre-trained weights, denoted as $\theta^\mu$, $\theta^D$, and $\theta^Q$ for the policy networks, discriminators, and Q-networks, respectively.

The training process uses several hyper-parameters, including the learning rate $\alpha$, gradient penalty coefficients $\lambda=\{\lambda_{vx}, \lambda_{vy}, \lambda_{rrx}, \lambda_{rry}, \lambda_{rcx}, \lambda_{rcy}\}$, and mini-batch size $m$. Additionally, the function $\mathbf{I}(\cdot)$ is defined such that $\mathbf{I}(True):=1$ and $\mathbf{I}(False):=0$ to aid in the computation of various loss functions and policy updates.

\section{Training Scheme}
\label{sec:alg}

The RL-V2V-GAN training algorithm integrates reinforcement learning and generative adversarial networks to achieve high-quality video-to-video synthesis. The training algorithm is shown as Algorithm \ref{alg_main} and visually depicted in Figure \ref{fig:modeL_RL}. The training process is designed to optimize the policy networks and discriminators through a combination of policy gradient and adversarial training. At a high level, the algorithm begins by initializing the replay buffer and network parameters, including the policy networks, Q-networks, and discriminators. During each iteration of the training loop, the algorithm processes each video in the dataset to construct transition mini-batches. This involves selecting sequences of frames as states, generating actions using the policy networks, and calculating rewards based on the adversarial, recurrent, recycle, and video losses. The transitions are stored in the replay buffer. The deep deterministic policy gradient (DDPG) method is then used to update the Q-networks and policy networks. Simultaneously, random batch of samples from the original dataset are used to update the discriminators based on the total losses of the GAN, ensuring both temporal coherence and stylistic consistency in the generated videos.

The DDPG approach relies on the expected finite horizon undiscounted return, denoted by $J$, which is represented as the cumulative reward expected from following the policy over a finite time horizon in this algorithm. The algorithm optimizes $J$ by computing the policy gradient, ensuring that the policy networks generate actions that maximize this expected return. This leads to effective updates of the networks, enabling the generation of high-quality video sequences.

Let $P \in \{P_x, P_y\}$ and $G, G' \in \{G_x, G_y\}$. The notation $G'(P(G(s)))$ requires special attention. Specifically, $G'$ indicates that the generator used in the second step must be different from the generator used in the first step. This is because the initial state $s$, which is a sequence of frames, must belong to either the source or target domain. The process involves first using one generator $G$ to transfer the state $s$ to the other domain, then using the corresponding predictor $P$ in the new domain to increment on the time axis, and finally using a different generator $G'$ to convert the state back to the original domain. This ensures that the transitions maintain domain consistency while capturing temporal dynamics.

The parameters $\sigma_1$ and $\sigma_2$ control the balance between exploration and exploitation. We set both $\sigma_1$ and $\sigma_2$ to constant values of 0.8. However, it is also a common practice to adjust these probabilities dynamically, allowing them to converge towards 1 or 0 as the training progresses. This convergence helps the model focus more on exploitation as it becomes more confident in its learned policies. We use the notation $\text{Bernoulli}(\sigma)$, where $\sigma \in \{\sigma_1, \sigma_2\}$. Sampling from $\text{Bernoulli}(\sigma)$ yields 1 with probability $\sigma$ and 0 with probability $1 - \sigma$. In the algorithm, $\sigma_1$ determines whether to load a transition from the replay buffer or to generate a new transition, while $\sigma_2$ decides whether to select the next action based on the maximum Q-value or choose a random action. The discount factor \( \gamma \) plays a critical role in balancing short-term and long-term rewards. A higher \( \gamma \) encourages the model to prioritize future gains, whereas a lower \( \gamma \) results in a strategy that favors immediate rewards. This balance allows the agent to plan ahead while still valuing the immediate outcomes of its actions.

Overall, the RL-V2V-GAN algorithm is designed to leverage the strengths of both reinforcement learning and GANs, ensuring that the generated videos are both temporally coherent and stylistically consistent with the target domain. The detailed steps and considerations discussed here are critical for achieving the high performance demonstrated by the model.

\vskip -0.1in
\begin{figure*}[htbp] 
\begin{minipage}{\textwidth}
\begin{algorithm}[H]
\normalsize
\caption{RL-V2V-GAN Training Algorithm}\label{alg_main}
 \hspace*{\algorithmicindent} \textbf{Input} The discriminators $D_x(s, a|\theta^{D_x}), D_y(s, a|\theta^{D_y})$, Q-networks $Q_x(s, a|\theta^{Q_x}), Q_y(s, a|\theta^{Q_y})$ and policy networks ${G_x}(s|\theta^{G_x})$, ${G_y}(s|\theta^{G_y})$, ${P_x}(s|\theta^{P_x})$ and ${P_y}(s|\theta^{P_y})$, learning rate $\alpha$, gradient penalty coefficients $\lambda=\{\lambda_{v}, \lambda_{rr}, \lambda_{rc}\}$, mini-batch size $m$, discount factor $\gamma$.\\
 \hspace*{\algorithmicindent} \textbf{Output} Trained policy networks $\mu(\cdot|\theta^\mu)$ and discriminators $D(\cdot|\theta^D)$
\begin{algorithmic}[1]
\State Initialize replay buffer $\mathcal{B}$.
\State Initialize $D$, $G$, and $\mu$ with pre-trained weights.
\State Initialize $Q'$ and $\mu'$ with $\theta^{Q'} = \theta^Q$ and $\theta^{\mu'} = \theta^{\mu}$.
\State \textbf{Loop} max epochs times:
\Indent
    \For{each video $v$ in training set $\mathcal{X} \cup \mathcal{Y} \cup \mathcal{Z}$}
        \State Set mini-batch $\textbf{B} = \emptyset$
        \State \textbf{Repeat} $m$ times:
            \Indent
            \State Sample $u_1 \sim \text{Bernoulli}(\sigma_1)$ to decide how to get the next transition
            \If{$u_1 = 1$ and $\mathcal{B} \neq \emptyset$}
                \State Load random tuple $(s, a, r, s', a^{\text{true}}, p)$ from reply buffer $\mathcal{B}$ and add to mini-batch $\textbf{B}$
            \Else
                \State Randomly pick $p \in \{1,2,\ldots,T-1\}$ and select $p$ consecutive frames from video $v$ as state $s$
                \State Use policy network to generate two candidate actions: $a_1 = P(s)$ and $a_2 = G^{(1)}(P(G^{(2)}(s)))$ where $P \in \{P_x, P_y\}, G^{(1)}, G^{(2)} \in \{G_x, G_y\}$, and the choice depends on the membership of $v$.
                \State Sample $u_2 \sim \text{Bernoulli}(\sigma_2)$ to decide how to choose the next action
                \If{$u_2 = 1$}
                    \State Select next action $a \in \arg \max \{Q(s,a_1), Q(s,a_2)\}$
                \Else
                    \State Select a random frame $a$ from $\boldsymbol{x} \cup \boldsymbol{y} \cup \boldsymbol{z} \cup \{\bar{y}\}$ as next action $a$
                \EndIf
                \If{$p = T-1$}
                    \State Obtain reward from video loss $r \leftarrow - \lambda_{v} L_v$ 
                \Else 
                    \State Obtain reward from adversarial, recurrent and recycle loss $r \leftarrow  - (L_g + \lambda_{rr} L_{rr} + \lambda_{rc} L_{rc})$ 
                \EndIf
                \State Get next state $s'$ by concatenating current state $s$ with the action $a$.
                \State Get ground truth action $a^{true}$ based on video $v$
                \State Add tuple $(s, a, r, s', a^{\text{true}}, p+1)$ to buffer $\mathcal{B}$ and to mini-batch $\textbf{B}$
            \EndIf
            \EndIndent
        \For{each transition $x = $ $(s, a, r, s', a^{\text{true}}, p) \in \textbf{B}$}
            \vspace{0.2em}
            \State Compute target: $ \hat{r}_{x} \leftarrow r + \gamma Q_{x}(s, \mu(s|\theta^{\mu'}) |\theta^{Q'_x}) \cdot \mathbf{I}(a^{\text{true}} \in \{\boldsymbol{x}\}) + \gamma Q_{y}(s, \mu(s|\theta^{\mu'}) |\theta^{Q'_y}) \cdot \mathbf{I}(a^{\text{true}} \in \{\boldsymbol{y}\}) $
            \vspace{0.2em}
        \EndFor
        \State Update $\theta^Q$ by minimizing the loss $\mathcal{L}^Q = \frac{1}{m} \sum_{x = (\bar{s}, \bar{a}, \cdot) \in \textbf{B}} (\hat{r}_{x} - Q(\bar{s}, \bar{a}|\theta^Q))^2$
        \vspace{0.2em}
        \State Update $\theta^\mu$ with deep deterministic policy gradient:
        \vspace{0.2em}
        \State \quad \quad $\nabla_{\theta^\mu}J \approx - \frac{1}{m} \sum_{(\bar{s}, \cdot) \in \textbf{B}} \nabla_a Q(s,a|\theta^Q)|_{s=\bar{s}, a=\mu(\bar{s})} \circ \nabla_{\theta^\mu} \mu(s|\theta^\mu)|_{s=\bar{s}}$
        \vspace{0.2em}
        \State Update target networks $\theta^{Q'} \leftarrow \tau \theta^Q + (1-\tau) \theta^{Q'} $ and $ \theta^{\mu'} \leftarrow \tau \theta^{\mu} + (1-\tau) \theta^{\mu'}$
        \State Construct batch $\hat{\textbf{B}}$ of size $m$ from $\boldsymbol{x}$ or $\boldsymbol{y}$ or $\boldsymbol{z}$, mimicking the membership of $v$ in $\mathcal{X} \cup \mathcal{Y} \cup \mathcal{Z}$. Videos in $\hat{\textbf{B}}$ are of type $\boldsymbol{x}_{:t}$ or $\boldsymbol{y}_{:t}$ or $\boldsymbol{z}_{:t}$ for randomly selected values of $t$
        \vskip 0.3em
        \State Let $\hat{\textbf{B}}_1$ be the set of those videos in $\hat{\textbf{B}}$ with $t = T$, $\hat{\textbf{B}}_2 = \hat{\textbf{B}} \setminus \hat{\textbf{B}}_1$
        \vskip 0.3em
        \State Update $\theta^D$ by gradient $\dfrac{\partial L}{\partial \theta^D}$ with respect to $\hat{\textbf{B}}_1$
        \vskip 0.5em
        \State Update $\theta^D$ by gradient $\dfrac{\partial (L - L_v)}{\partial \theta^D}$ with respect to $\hat{\textbf{B}}_2$
    \EndFor
\EndIndent
\end{algorithmic}
\end{algorithm}
\end{minipage}
\end{figure*}

\vskip -0.1in
\vskip 3in

\section{Numerical Experiments}
\label{sec:v2v:experiments}
In this section, we present our experimental results,
which demonstrates the effectiveness of our proposed method
in terms of video generation quality compared to the state-of-the-art methods.

\subsection{Datasets}

We performed experiments on four distinct datasets to assess the performance of our proposed method for unsupervised video-to-video translation. These datasets provide comprehensive benchmarks for evaluating the efficacy of our approach. 

The first dataset consists of synthetic videos. The $\mathcal{X}$ set contains 600 frames of colorful rectangles moving on a black background, while $\mathcal{Y}$ and $\mathcal{Z}$ have 100 and 100 frames of circles moving on color-changing and black backgrounds, respectively. The 100 supplemental style images $\bar{\mathcal{Z}}$ consist of random circles on either blue or red backgrounds. Samples are shown in Table \ref{table-datasets-artificial}.

The second dataset, a flower dataset, contains 550 frames of red flowers blooming in black backgrounds as $\mathcal{X}$, 100 frames of yellow flowers blooming in natural backgrounds as $\mathcal{Y}$, 900 images of yellow flowers blooming in black backgrounds as $\mathcal{Z}$, and 2,000 images of random color flowers on natural backgrounds as $\bar{\mathcal{Z}}$. The $\mathcal{X}$ and $\mathcal{Z}$ videos are from the StyleGAN flower dataset, while the $\mathcal{Y}$ and $\bar{\mathcal{Z}}$ sets are from the YouTube-8M dataset. Samples are shown in Table \ref{table-datasets-flower}.

The third dataset, a city aerial dataset, contains 2,000 frames of daytime modern city aerial videos as $\mathcal{X}$, 500 frames of night time small-town aerials as $yv^{T}$, 1,500 images of nighttime city aerial photos as $\bar{\mathcal{Z}}$, and 1,000 frames of small-town aerial videos during daytime as $\mathcal{Z}$. All of the videos are from the YouTube-8M dataset. Samples are shown in Table \ref{table-datasets-city}.

The fourth, proprietary dataset, encompasses both synthetic and real-world images of 2 different types. It consists of 2,000 synthetic images labeled $\mathcal{X}$, alongside 500 real-world images of type A, tagged as $\mathcal{Y}$. Additionally, there are 1,000 real-world images of type B, categorized under $\mathcal{Z}$, and 4,000 real-world images depicting type A, identified as $\bar{\mathcal{Z}}$. For an in-depth overview of the dataset's composition, see Tables \ref{table-datasets-corporate} and \ref{table-datasets-corporate-2}.

\begin{table*}
\caption{Data examples of the artificial dataset}
\label{table-datasets-artificial}
\begin{center}
\begin{small}
\begin{sc}
\begin{tabular}{ccc}
\toprule
       & Source Domain & Target Domain \\
 Description & \makecell{rectangles \\ moving in black background}
 & \makecell{ circles \\ moving in changing backgrounds} \\ 
 \midrule
 & \includegraphics[width=0.4\linewidth]{figs_v2v/artificial_video_A}  & \includegraphics[width=0.4\linewidth]{figs_v2v/artificial_video_B} \\
 Video Data  & \makecell{ $\mathcal{X}$, 600 frames of colorful rectangles}  & $\mathcal{Y}$, 100 frames, colorful circles  \\
 & & \includegraphics[width=0.4\linewidth]{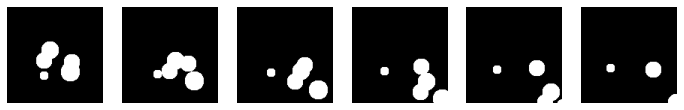} \\ 
 & & $\mathcal{Z}$, 100 frames, gray circles \\
  \midrule
  Image Data & None & \includegraphics[width=0.4\linewidth]{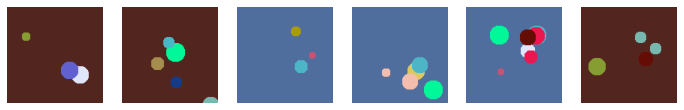}  \\ 
  &  & \makecell{$\bar{\mathcal{Z}}$, 100 images, color circles}  \\ 
 \bottomrule
\end{tabular}
\end{sc}
\end{small}
\end{center}
\vskip -0.1in
\end{table*}

\begin{table*}
\caption{Data examples of the flower dataset}
\label{table-datasets-flower}
\begin{center}
\begin{small}
\begin{sc}
\begin{tabular}{ccc}
\toprule
       & Source Domain & Target Domain \\
 Description & \makecell{red flowers \\ black background}
 & \makecell{ yellow flower \\ in nature} \\ 
 \midrule
 Video Data & \includegraphics[width=0.4\linewidth]{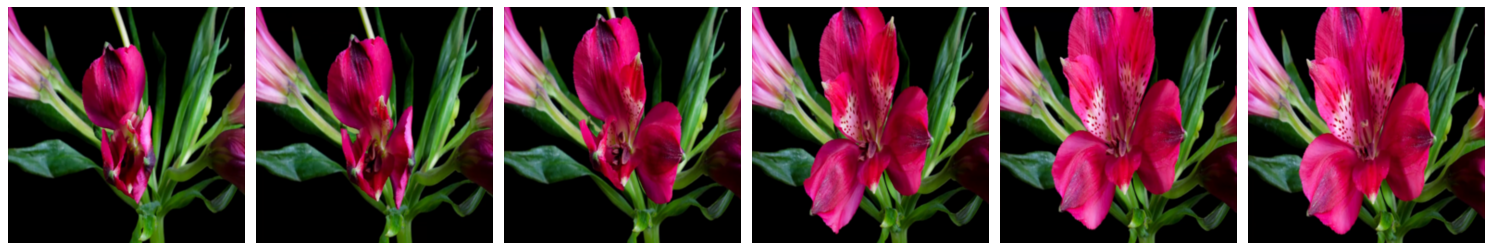}  & \includegraphics[width=0.4\linewidth]{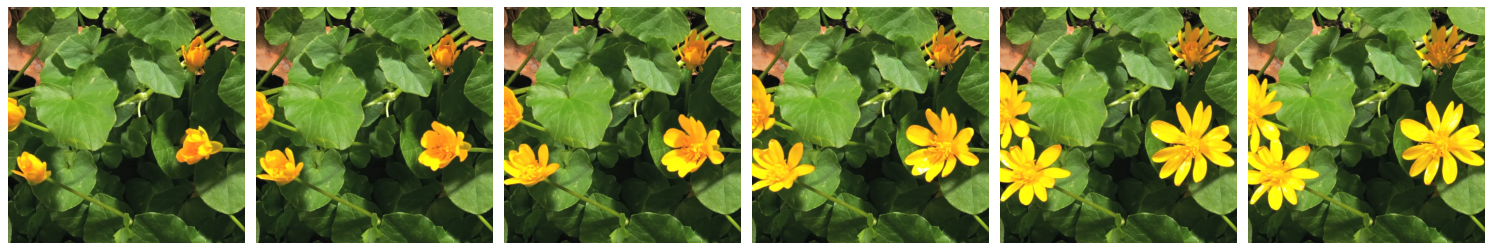} \\
   & \makecell{ $\mathcal{X}$, 550 frames}  & $\mathcal{Y}$, 100 frames, in nature  \\
   & & \includegraphics[width=0.4\linewidth]{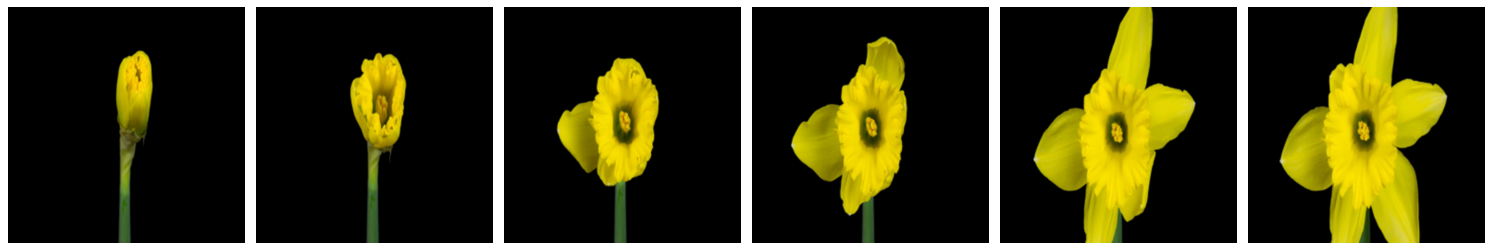} \\ 
 & & $\mathcal{Z}$, 900 frames, in black \\
  \midrule
  Image Data & None & \includegraphics[width=0.4\linewidth]{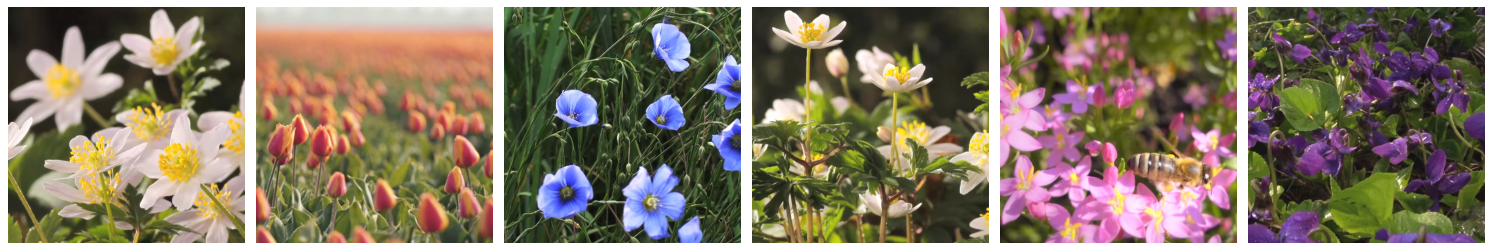}  \\ 
  &  & \makecell{$\bar{\mathcal{Z}}$, 2000 images, in nature}  \\ 
 \bottomrule
\end{tabular}
\end{sc}
\end{small}
\end{center}
\vskip -0.1in
\end{table*}

\begin{table*}
\caption{Data examples of the city aerial dataset}
\label{table-datasets-city}
\vskip 0.15in
\begin{center}
\begin{small}
\begin{sc}
\begin{tabular}{ccc}
\toprule
       & Source Domain & Target Domain \\
 Description & \makecell{modern cities \\ in daytime}
 & \makecell{ small-town \\ in nighttime} \\ 
 \midrule
 & \includegraphics[width=0.4\linewidth]{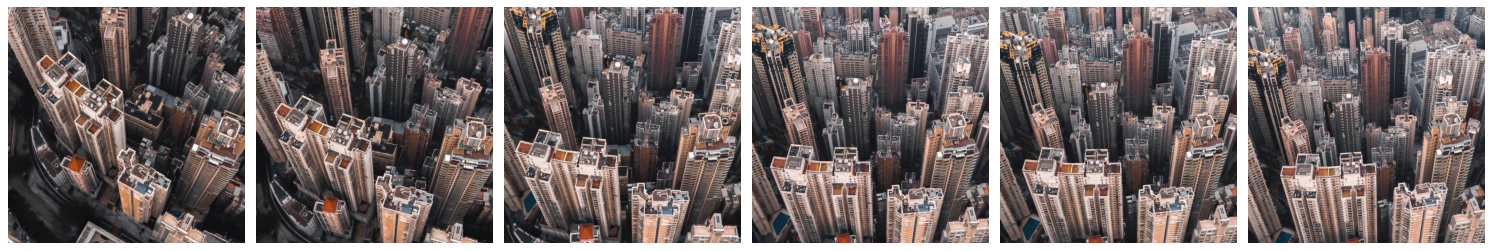}  & \includegraphics[width=0.4\linewidth]{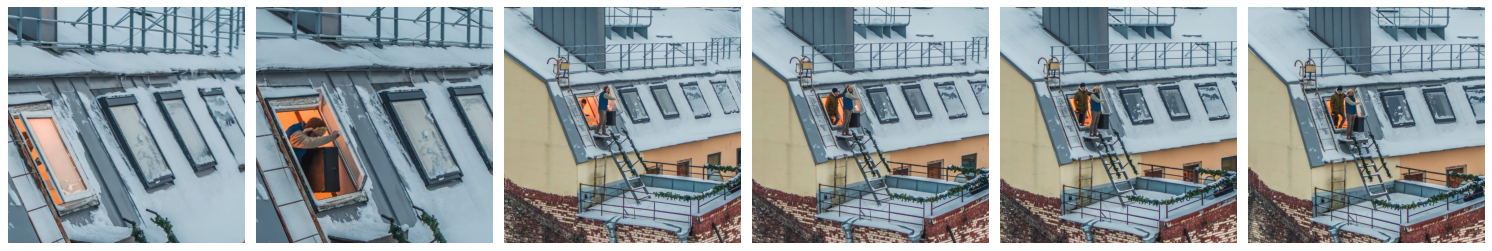} \\
 Video Data  & \makecell{ $\mathcal{X}$, 2000 frames}  & $\mathcal{Y}$, 500 frames, small-town nighttime  \\
 & & \includegraphics[width=0.4\linewidth]{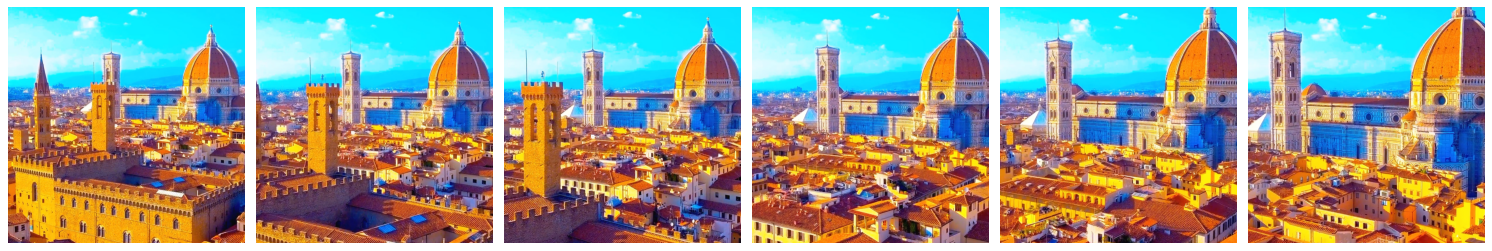} \\ 
 & & $\mathcal{Z}$, 1000 frames, small-town day-time \\
  \midrule
  Image Data & None & \includegraphics[width=0.4\linewidth]{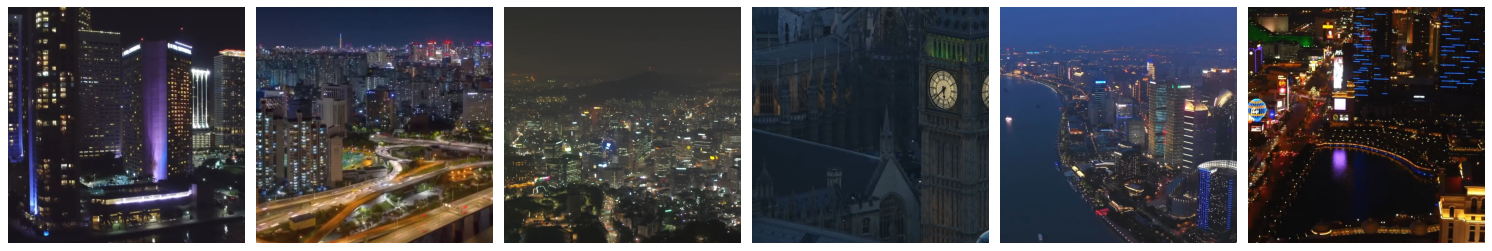}  \\ 
  &  & \makecell{$\bar{\mathcal{Z}}$, 1500 images, modern city nighttime}  \\ 
 \bottomrule
\end{tabular}
\end{sc}
\end{small}
\end{center}
\vskip -0.1in
\end{table*}

\begin{table}[htbp]
\caption{Description of the proprietary dataset source domain}
\label{table-datasets-corporate}
\begin{center}
\begin{small}
\begin{sc}
\begin{tabular}{cc}
\toprule
       & Source Domain \\
 Description & Synthetic \\ 
\midrule
 Video Data & $\mathcal{X}$, type A, 2,000 frames \\ 
\midrule
 Image Data  & None \\
 \bottomrule
\end{tabular}
\end{sc}
\end{small}
\end{center}
\vskip -0.1in
\end{table}

\begin{table}[htbp]
\caption{Description of the proprietary dataset target domain}
\label{table-datasets-corporate-2}
\begin{center}
\begin{small}
\begin{sc}
\begin{tabular}{cc}
\toprule
       &  Target Domain \\
 Description & Real World \\ 
\midrule
 Video Data & \makecell{$\mathcal{Y}$, type A, 500 frames \\
 $\mathcal{Z}$, type B, 1,000 frames} \\
\midrule
 Image Data  & $\bar{\mathcal{Z}}$, type A, 4,000 images \\
 \bottomrule
\end{tabular}
\end{sc}
\end{small}
\end{center}
\vskip -0.1in
\end{table}

\begin{table}[htbp]
\caption{Generated Samples from RL-V2V-GAN}
\label{fig:generated-samples}
\begin{center}
\begin{small}
\begin{sc}
\begin{tabular}{c}
Artificial Dataset \\
\includegraphics[width=0.95\linewidth]{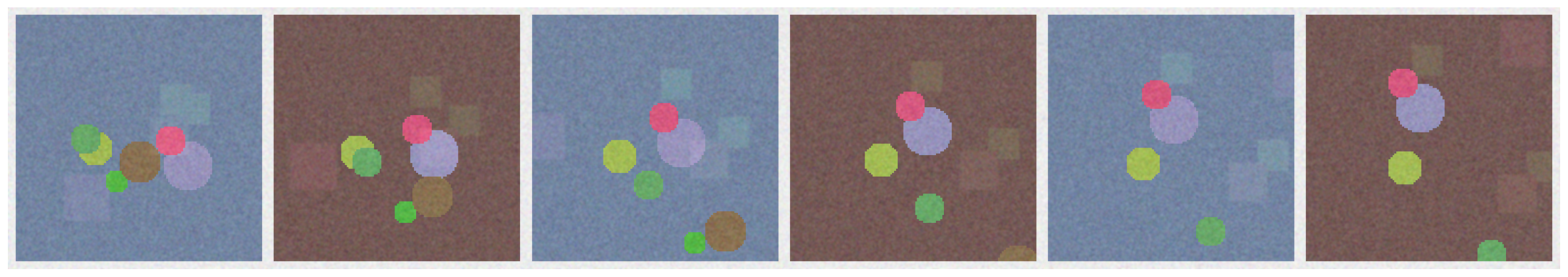} \\
\includegraphics[width=0.95\linewidth]{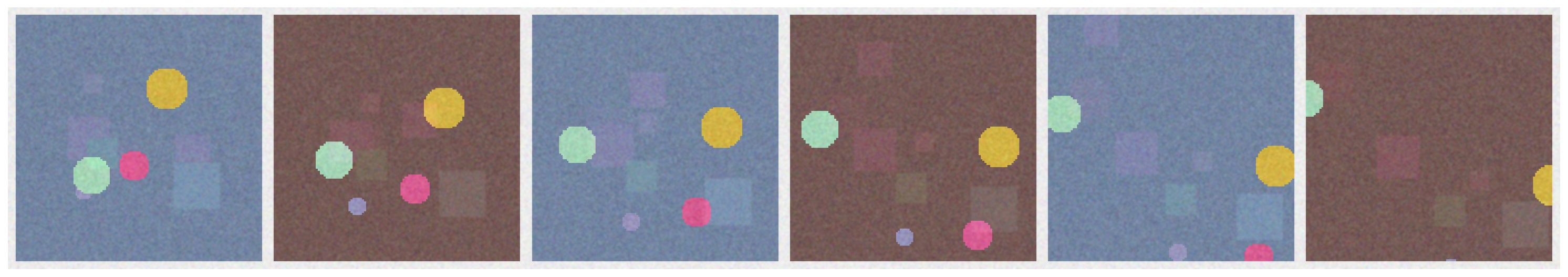} \\

Flower Dataset \\
\includegraphics[width=0.95\linewidth]{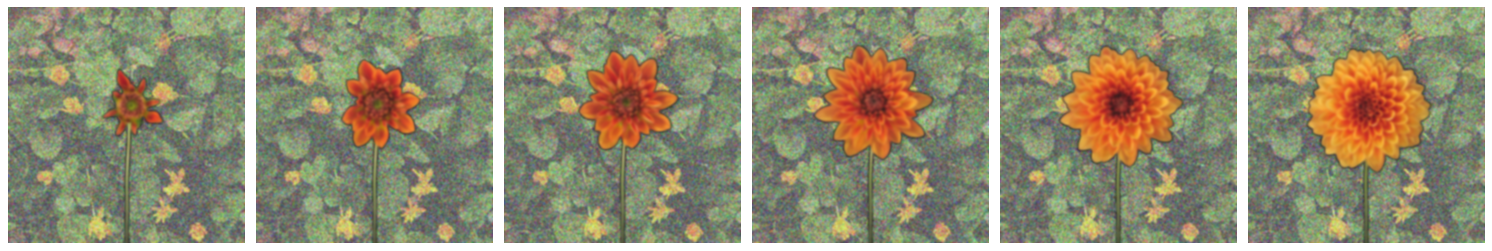} \\
\includegraphics[width=0.95\linewidth]{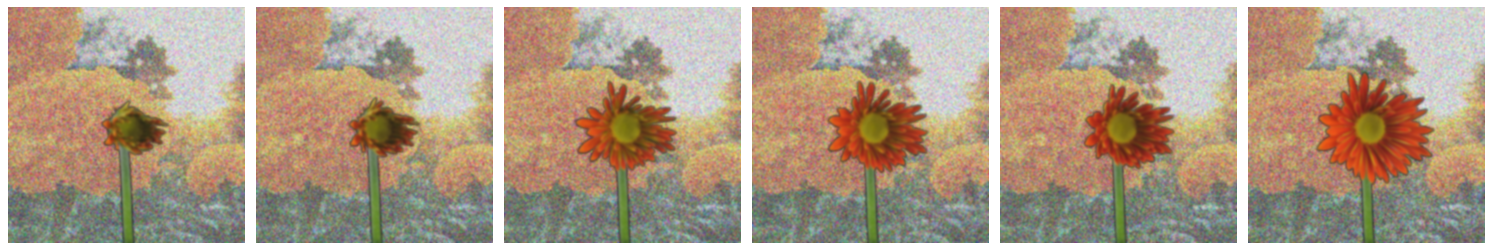} \\

City Aerial Dataset \\
\includegraphics[width=0.95\linewidth]{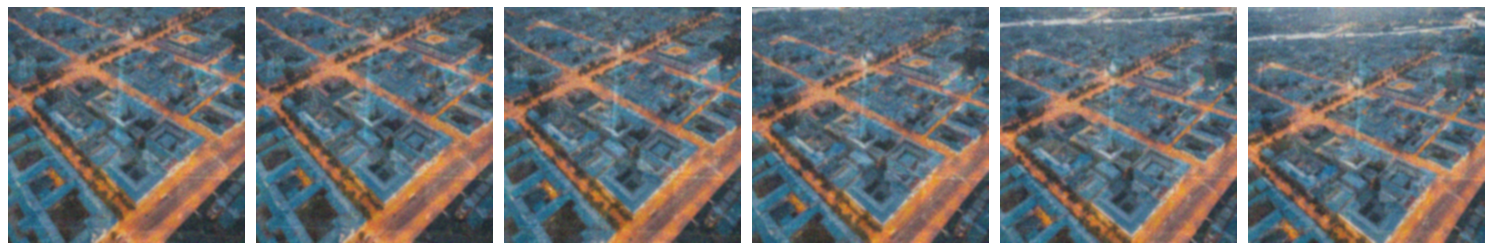} \\
\includegraphics[width=0.95\linewidth]{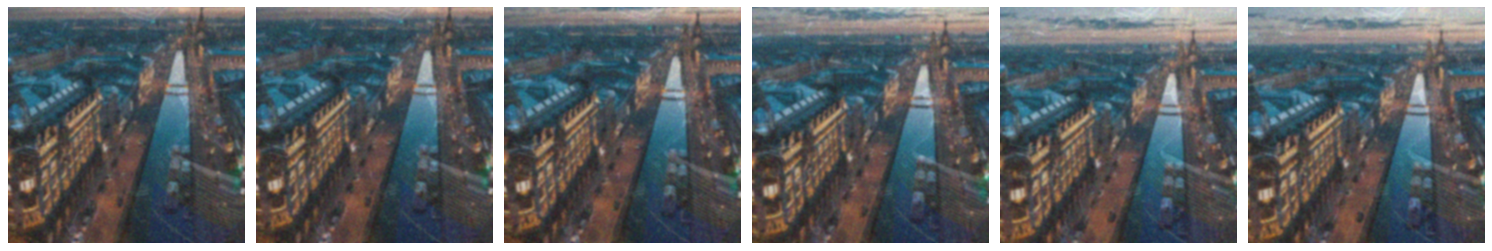} \\
\end{tabular}
\end{sc}
\end{small}
\end{center}
\vskip -0.1in
\end{table}

\subsection{Implementation}

We implemented the model using the Tensorflow 1.15 framework and trained it on four NVIDIA 3070 GPUs or equivalents. We used the softmax loss function for the auto-encoder and applied L2 regularization of penalty ratio of 0.001 to the model's trainable parameters. These parameters were initialized using Xavier initialization.

For the seq2seq auto-encoder, sequence-wise normalization was applied across multiple video sequences within each mini-batch. We computed the mean and variance statistics across all timesteps within this mini-batch for each output channel. Activation functions are ReLU. The optimization of our proposed model was performed using the stochastic gradient descent optimizer with a batch size of 64. Training was executed over 100 epochs, with an early stopping strategy that halts training if the validation set FID score  \cite{karras_style-based_2019} does not decrease for 5 consecutive epochs to prevent overfitting. We initially set the learning rate at $0.0001$, applying a decaying schedule that reduces it by a factor of 10 every 10 epochs to refine parameter adjustments. The optimizer configuration included a weight decay of $3 \cdot 10^{-3}$ and a momentum of $0.97$, optimizing the convergence process. We use gradient clipping. To update the target networks, we use soft updates with $\tau$ to control the degree of the update. Specifically, $\tau$ determines the interpolation between the current parameters of the target networks and the learned parameters. In our experiments, we chose $\tau = 0.005$.

The encoder contained 3 RPB layers and a dense layer. Each RPB block took an input tensor of size 256 x 256 x 3. The ResConvLSTM layer employed a 3 x 3 kernel, a stride of 1, and had 64 hidden states. The subsequent ResConvLSTM used the same kernel size, padding, and stride but contained 32 hidden states. The output tensor after the residual connection was of size 256 x 256 x 16, which was then reduced to 128 x 128 x 16 after pooling. The output tensor was further reduced to 32 x 32 x 16 after the third RPB block. This tensor was mapped to a 4,000-entry embedding vector by a dense layer. 

For the decoders, the URB block's embedding vector was first transformed by a dense layer to 32 x 32 x 16 before entering the first URB block. Each URB block's ResConvLSTM utilized a 3 x 3 kernel, a stride of 1, and had 32 hidden states. The final ConvLSTM layer in the decoder had 3 filters, producing an output tensor of size 256 x 256 x 3. The discriminators and Q-networks have two last, fully-connected layers of 1,000 and 200 neurons. 

In our experiments, the values of the lambda parameters $\lambda_{rrx}$, $\lambda_{rry}$, $\lambda_{rcx}$, $\lambda_{rcy}$, $\lambda_{vx}$, and $\lambda_{vy}$ were selected based on a combination of empirical tuning and theoretical considerations. For the artificial and proprietary datasets, we set $\lambda_{rrx}$ and $\lambda_{rry}$ to 1 to ensure strong temporal coherence in the predicted frames. The values of $\lambda_{rcx}$ and $\lambda_{rcy}$ were set to 0.5 to balance the ReCycle losses, promoting consistency in style and domain transitions. For the video loss terms, $\lambda_{vx}$ and $\lambda_{vy}$ were set to 1, giving a moderate weight to overall video quality while allowing adversarial and recurrent losses to dominate. For the flower dataset, we used $\lambda_{rrx}$ and $\lambda_{rry}$ values of 1, $\lambda_{rcx}$ and $\lambda_{rcy}$ values of 0.125, and $\lambda_{vx}$ and $\lambda_{vy}$ values of 0.5. For the city aerial dataset, we used $\lambda_{rrx}$ and $\lambda_{rry}$ values of 1, $\lambda_{rcx}$ and $\lambda_{rcy}$ values of 0.5, and $\lambda_{vx}$ and $\lambda_{vy}$ values of 0.5. These values were determined through grid search to optimize the validation set FID scores. In our numerical experiments, the discount factor $\gamma$ is set to 0.99 to prioritize long-term rewards and account for the sequential nature of video frames.

The raw videos varied in length and resolution. We normalized them to 30 frames per second, cropped to the center 256 x 256 squares, and standardized the values on each channel to have be mean zero and unit variance across all videos. This standardized the data in terms of frame size, rate, and value ranges to effectively train deep learning models on the dataset. For the purpose of our numerical experiments, we set the video length to $T=6$, as this allows the model to focus on key transitions and avoid overfitting to long, potentially redundant sequences, while still capturing essential temporal dynamics efficiently. Longer sequences can be supported by the algorithm as needed. All hyper parameters were either set based on experience or were partially ad-hoc optimized. In our numerical experiments, the capacity of reply buffer $\mathcal{B}$ is $10,000$.

The training process involves two key phases: pre-training and reinforcement learning fine-tuning. In the pre-training phase, the autoencoder behind generators $G_x$, $G_y$, and predictors $P_x$ and $P_y$ are trained from scratch using adversarial loss, recurrent loss, and recycle loss. The autoencoder is trained for up to 50 epochs, with early stopping based on the FID score from the validation set. After pre-training convergence, we introduce the proposed video loss function and continue training the model parameters using reinforcement learning. The pre-trained model is fine-tuned for 100 additional epochs or until early stopping criteria are met. 

\subsection{Benchmark and Evaluation}

In this study, we evaluated the performance of our proposed model against two state-of-the-art algorithms, RecycleGAN and MoCoGAN, on various benchmark datasets. RecycleGAN is an unsupervised video retargeting that seamlessly integrates spatial and temporal information with adversarial and recycle losses. MoCoGAN, on the other hand, is a motion and content decomposed generative adversarial network framework for video generation. It maps a sequence of random vectors to a sequence of video frames, with each vector consisting of a content and motion part, and is capable of generating videos with interchangeable content and motion.

In our evaluation, we use the Fréchet Inception Distance (FID) to assess our model's performance, as it comprehensively evaluates the quality of generated videos by comparing feature vectors from generated and real videos using a pre-trained Inception v3 network. FID's ability to measure differences in distribution means and covariances has made it the preferred metric, overcoming limitations of other measures. The lower the FID, the better the model's performance.

\subsection{Results}

\begin{table}[htbp]
\caption{FID Score for generated samples from all datasets.}
\label{numerical-results}
\begin{center}
\begin{scriptsize}
\begin{sc}
\begin{tabular}{lcccr}
\toprule
Data set & ReCycleGAN & MoCoGAN & RL-V2V-GAN \\
\midrule
Artificial    & 9.9 & 9.7 & 6.2 \\
Proprietary     & 8.3 & 8.0 & 7.3 \\
Flower        & 6.9 & 8.8 & 6.9 \\
City Aerial   & 7.8 & 8.3 & 7.5 \\
\bottomrule
\end{tabular}
\end{sc}
\end{scriptsize}
\end{center}
\end{table}

Our results, as shown in Table \ref{numerical-results}, indicate that RL-V2V-GAN performs the best among the three models on the artificial, proprietary, and aerial datasets, with FID scores of 6.2, 7.3, and 7.5 respectively. Meanwhile, on the flower dataset, both RL-V2V-GAN and ReCycleGAN perform similarly with an FID score of 6.9. In general, the results demonstrate the superiority of our proposed RL-V2V-GAN model, which consistently outperforms the benchmarks and has a lower FID score, indicating higher-quality results. These findings validate the effectiveness of our proposed method in transforming videos from one domain to another in an unsupervised manner. For inference, the dataset is split 70/15/15 into the train, validation, and test sets. The test set is evaluated with the FID score.

\begin{table}[htbp]
\caption{Computational time for each model}
\label{computational-time}
\begin{center}
\begin{scriptsize}
\begin{sc}
\resizebox{0.46 \textwidth}{!}{
\begin{tabular}{lcccccc}
\toprule
Data set & Model & \makecell{Runtime \\ per \\ epoch \\ (min)} & \makecell{Epochs \\ to \\ converge} & \makecell{Runtime \\ to \\ converge \\ (min)} \\
\midrule
\multirow{3}{*}{Artificial} & MoCoGAN     & \phantom{0}2 & 20 & \phantom{00}40 \\
                            & ReCycleGAN  & \phantom{0}3 & 27 & \phantom{00}81 \\
                            & RL-V2V-GAN  & \phantom{0}4 & \phantom{0}6  & \phantom{00}24 \\
\midrule
\multirow{3}{*}{Proprietary}  & MoCoGAN     & 24 & 25 & \phantom{0}600 \\
                            & ReCycleGAN  & 35 & 31 & 1085 \\
                            & RL-V2V-GAN  & 47 & 16  & \phantom{0}752 \\
\midrule
\multirow{3}{*}{Flower}     & MoCoGAN     & 13 & 35  & \phantom{0}455 \\
                            & ReCycleGAN  & 20 & 44  & \phantom{0}880 \\
                            & RL-V2V-GAN  & 25 & 23  & \phantom{0}575 \\
\midrule
\multirow{3}{*}{City Aerial}& MoCoGAN     & 18 & 32 & \phantom{0}576 \\
                            & ReCycleGAN  & 27 & 40 & 1080 \\
                            & RL-V2V-GAN  & 35 & 18  & \phantom{0}630 \\
\bottomrule
\end{tabular}
}
\end{sc}
\end{scriptsize}
\end{center}
\end{table}

Table \ref{computational-time} presents the computational time for each model across four different datasets. The results indicate that while the RL-V2V-GAN model generally requires more time per epoch compared to MoCoGAN and ReCycleGAN, it significantly reduces the total runtime to convergence due to requiring fewer epochs. For instance, in the artificial dataset, RL-V2V-GAN converges in only 24 minutes, compared to 40 minutes for MoCoGAN and 81 minutes for ReCycleGAN. Similarly, for the proprietary dataset, RL-V2V-GAN converges in 752 minutes, whereas MoCoGAN and ReCycleGAN take 600 and 1085 minutes, respectively. This trend is consistent across all datasets, demonstrating that RL-V2V-GAN's efficient learning process results in quicker convergence. The proposed model's ability to achieve high-quality results in fewer epochs highlights its superior performance and efficiency in video-to-video synthesis tasks.

\begin{figure}[htbp]
    \centering
    \begin{subfigure}[b]{0.495\linewidth}
        \centering
        \includegraphics[width=\linewidth]{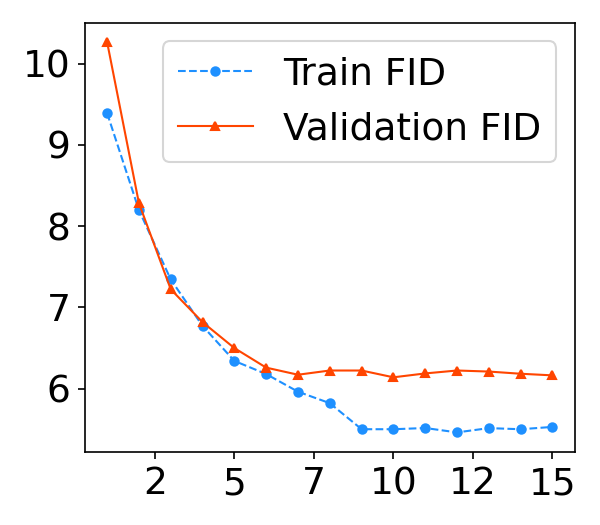}
        \vskip -0.1in
        \caption{Artificial Dataset}
        \label{fig:artificial_dataset}
    \end{subfigure}
    \hfill
    \begin{subfigure}[b]{0.495\linewidth}
        \centering
        \includegraphics[width=\linewidth]{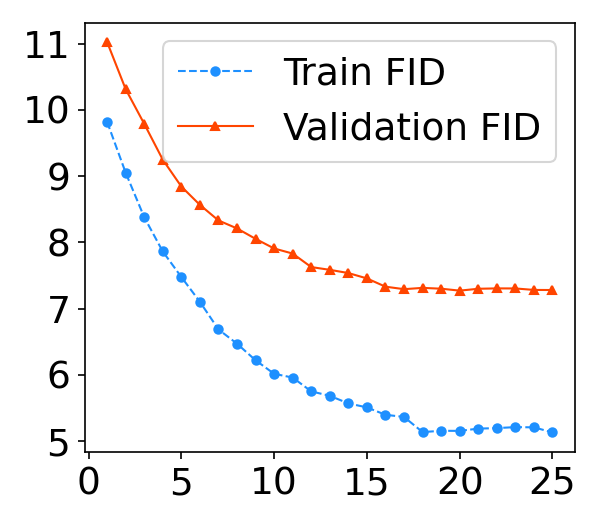}
        \vskip -0.1in
        \caption{Proprietary Dataset}
        \label{fig:corporate_dataset}
    \end{subfigure}
    \vskip 0.1in
    \begin{subfigure}[b]{0.495\linewidth}
        \centering
        \includegraphics[width=\linewidth]{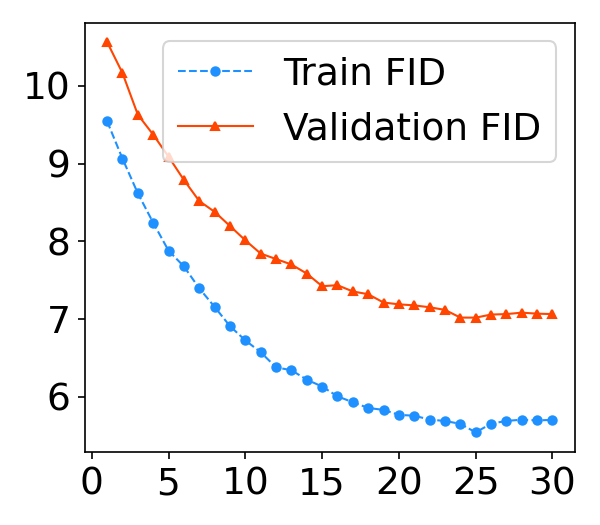}
        \vskip -0.1in
        \caption{Flower Dataset}
        \label{fig:flower_dataset}
    \end{subfigure}
    \hfill
    \begin{subfigure}[b]{0.495\linewidth}
        \centering
        \includegraphics[width=\linewidth]{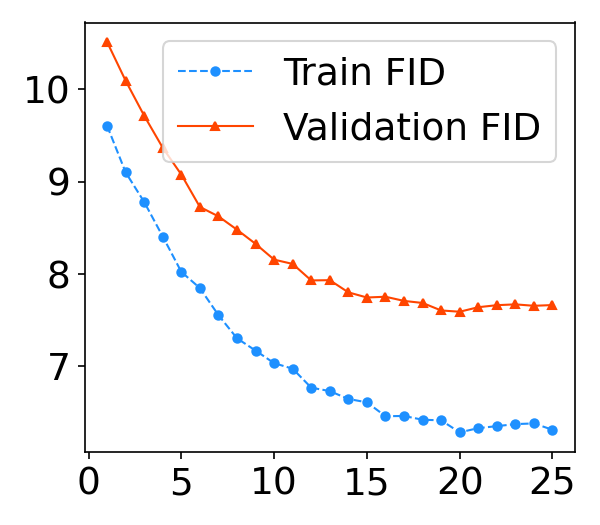}
        \vskip -0.1in
        \caption{City Aerial Dataset}
        \label{fig:city_aerial_dataset}
    \end{subfigure}
    \caption{Train and validation FID scores over reinforcement learning epochs for the Artificial, Proprietary, Flower, and City Aerial datasets. Each subplot shows the training (blue) and validation (orange) FID scores, highlighting the model's learning dynamics and generalization capabilities.}
    \label{fig:train_validation_fid}
\end{figure}

Figure \ref{fig:train_validation_fid} shows the training and validation FID scores as a function of reinforcement learning epochs for the Artificial, Proprietary, Flower, and City Aerial datasets. Across all datasets, the training FID scores (blue curves) decrease rapidly initially, indicating efficient learning. The validation FID scores (orange curves) generally follow a similar trend, suggesting good generalization. The consistent gap between training and validation FID scores in each dataset indicates that the model avoids overfitting. Particularly, the Artificial and Flower datasets show significant early improvements, while the Proprietary and City Aerial datasets demonstrate steady progress. These results highlight the RL-V2V-GAN model's ability to learn and maintain performance across diverse datasets, producing high-quality video synthesis.

\subsubsection{Reinforcement Learning Training Strategies}

\begin{figure}[htbp]
\begin{center}
\includegraphics[width=0.95\linewidth]{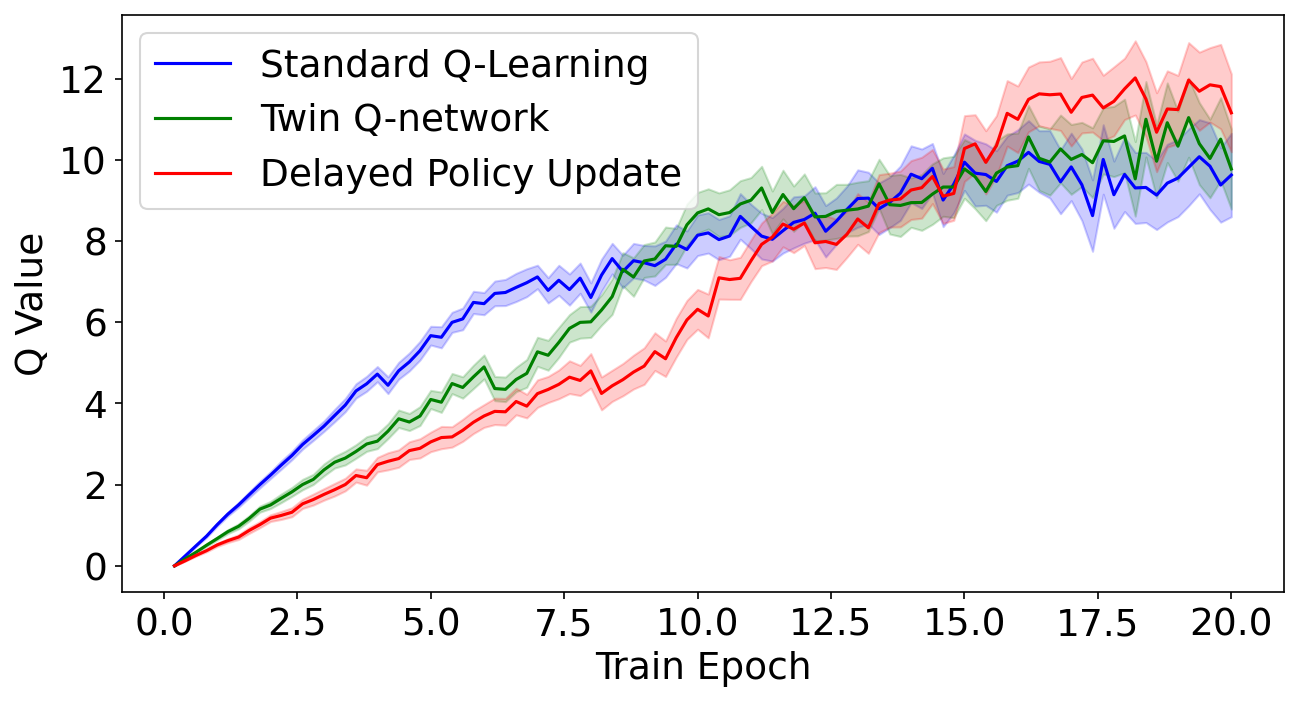} \\
\caption{Reward progression for standard Q-learning, twin Q-network, and delayed policy update. Shaded regions indicate the variability (standard deviation) across multiple (n=10) independent runs.}
\label{fig-results-compare-policy}
\end{center}
\vskip -0.2in
\end{figure}

In Figure \ref{fig-results-compare-policy}, we compare the performance of three distinct reinforcement learning training schemes: standard Q-learning, twin Q-network, and delayed policy update methods. This study is conducted on the city aerial dataset. The standard Q-learning approach updates the value of an action in a state by using the weighted average of the current value and the newly acquired information. Although this method showed promising results in the early epochs, it led to the lowest Q values upon convergence. The twin Q-network approach employs two separate Q-value estimators to provide unbiased Q-value estimates, achieving moderate Q values at both the initial and final epochs. Finally, the delayed policy update method integrates techniques from clipped double-Q learning and target policy smoothing. Initially, it displayed the lowest Q values, but it ultimately reached the highest Q values at convergence, illustrating its potential to enhance stability in traditional reinforcement learning training while decreasing the time required for convergence. We use the delayed policy update method in our main computation results due to its superior performance in achieving higher Q values and enhancing training stability.

\subsection{Ablation Studies}

We set up an ablation study centered on evaluating the efficacy of disparate training methodologies applied to our proposed model. The results in Table \ref{results-tricks} show the comparative impacts of pre-training across all four datasets. The first column in the table shows the results of training the model from scratch without any pre-training. The second column shows the results of using pre-trained models. The results show that pre-training the model can significantly improve its performance, as demonstrated by the lower FID scores in the second column compared to the first.

\begin{table}[htbp]
\caption{FID score for comparing training from scratch vs using pre-trained weights.}
\label{results-tricks}
\begin{center}
\begin{scriptsize}
\begin{sc}
\begin{tabular}{lccc}
\toprule
Data set  & No Pre-train         & Pre-trained \\
\midrule
Artificial & 11.2 & 6.2 \\
Proprietary  &  \phantom{0}9.5 & 7.3 \\
Flower     & 10.6 & 6.9 \\
Aerial     & 11.9 & 7.5 \\
\bottomrule
\end{tabular}
\end{sc}
\end{scriptsize}
\end{center}
\vskip -0.1in
\end{table}

This ablation study demonstrates that pre-training is an effective technique for improving the performance of our proposed model. By providing better initial parameters, pre-training speeds up convergence and improves performance, as shown by the lower FID scores.

We further conducted an ablation study to investigate the impact of different loss components on the performance of our RL-V2V-GAN model. Specifically, we analyzed the effect of setting each type of loss — adversarial, recurrent, recycle, and video — to zero. The FID scores for three datasets (Artificial, City Aerial, and Flower) under all four scenarios are shown in Table \ref{table-ablation-lambda}.

\begin{table}[htbp]
\caption{FID scores for ablation study on the effect of removing different types of losses.}
\label{table-ablation-lambda}
\begin{center}
\begin{scriptsize}
\begin{sc}
\begin{tabular}{lccc}
\toprule
\multicolumn{1}{c}{} & \multicolumn{3}{c}{\textbf{Dataset}} \\
\cmidrule(r){2-4}
\textbf{Loss Type} & \textbf{Artificial} & \textbf{City Aerial} & \textbf{Flower} \\
\midrule
All Losses         & 6.2  & \phantom{0}7.5  & 6.9 \\
No Adversarial Loss& 8.4  & \phantom{0}8.9  & 8.1 \\
No Recurrent Loss  & 9.3  & 10.2 & 9.1 \\
No Recycle Loss    & 8.7  & 10.1 & 8.9 \\
No Video Loss      & 7.5  & \phantom{0}9.4  & 8.5 \\
\bottomrule
\end{tabular}
\end{sc}
\end{scriptsize}
\end{center}
\end{table}

The results indicate that the removal of any loss component leads to a deterioration in performance, as evidenced by higher FID scores across all datasets. The first row in the table represents the performance of the model when all four loss components — adversarial, recurrent, recycle , and video — are included, serving as the baseline for comparison. For the Artificial dataset, the FID scores increased from 6.2 to 8.4, 9.3, 8.7, and 7.5 when the adversarial, recurrent, recycle, and video losses were removed, respectively, demonstrating the importance of each loss component. Similarly, for the City Aerial dataset, FID scores rose from 7.5 to 8.9, 10.2, 10.1, and 9.4, and for the Flower dataset, from 6.9 to 8.1, 9.1, 8.9, and 8.5 for the removal of the same losses.

Overall, this ablation study confirms the necessity of each loss component in our RL-V2V-GAN model to ensure the generation of temporally coherent, stylistically consistent, and high-quality videos. However, the video loss component appears to have a slightly lesser impact on quality compared to the recurrent and recycle losses.

\section{Limitation and Future Directions}
The RL-V2V-GAN architecture, while demonstrating significant potential in the domain of video retargeting, is confronted with certain constraints that merit attention. A notable challenge lies in the model's current constraint of generating singular style outputs from a given input image. This limitation suggests an avenue for enhancement through the integration of a one-to-many translation model within the framework. Such an advancement would not only elevate the output's diversity but also its creative dimension, paving the way for a richer variety of video retargeting possibilities.

Moreover, the capability of the model to process a limited quantity of video frames emerges as a constraint, rooted in the extensive computational requirements necessitated by the high-dimensional nature of video data within the neural network architecture. Given the inherently multimodal character of videos, which frequently encompasses both visual and audio streams, this limitation underscores the complexity of video processing. A promising direction for future research involves the development of a multimodal model adept at assimilating various data inputs, including audio. This approach holds the potential to refine the model's output, fostering a more comprehensive and nuanced interpretation of the data, thereby enriching the resultant video retargeting outcomes.

\section{Conclusion}
This work introduces RL-V2V-GAN, a model in the realm of unsupervised video-to-video synthesis, leveraging the strength of reinforcement learning and GANs for the generation of temporally coherent video sequences in a target style from limited source material. Our approach effectively mitigates the challenge of sparse data in the target domain, demonstrating a significant advancement over existing methods through its ability to learn and replicate complex video styles without requiring paired input videos.

The experimental outcomes underscore the model's superior performance across diverse datasets, as evidenced by favorable FID scores when benchmarked against state-of-the-art models like ReCycleGAN and MoCoGAN. This success is attributed to our model's innovative integration of reinforcement learning, which enables nuanced temporal dynamics modeling, and a carefully designed GAN architecture that ensures style consistency and content fidelity.

RL-V2V-GAN's efficiency in synthesizing high-quality video content opens new avenues for applications in areas demanding robust video manipulation capabilities, such as content creation, multimedia editing, and virtual reality environments. By achieving a delicate balance between generative flexibility and output coherence, this work opens up new opportunities in the field of video generation and expands the current capabilities of generative models, paving the way for future research to explore even more sophisticated models within this promising domain.

\bibliography{references}
\bibliographystyle{icml2021}

\clearpage

\section*{Appendix: Notation Table}

\begin{table}[htbp]
\begin{center}
\begin{small}
\begin{sc}
\resizebox{0.99\textwidth}{!}{
\begin{tabular}{lll}
\toprule
Variable & Description & Dimension \\
\midrule
$N$ & Number of videos in a domain & (1) \\
$T$ & Number of frames in a full video & (1) \\
$W$ & Width of a frame & (1) \\
$H$ & Height of a frame & (1) \\
$\mathcal{X}$ & Source domain videos & (N, T, W, H) \\
$\mathcal{Y}$ & Target domain videos & (N, T, W, H) \\
$\mathcal{Z}$ & Target domain style videos & (N, T, W, H) \\
$\bar{\mathcal{Z}}$ & Target domain style images & (N, W, H) \\
$\boldsymbol{x}_{:t}\in \mathcal{X}$ & Video sequence from source domain & (T, W, H)\\
$\boldsymbol{y}_{:t}\in \mathcal{Y}$ & Video sequence from target domain & (T, W, H)\\
$\boldsymbol{z}_{:t}\in \mathcal{Z}$ & Style video sequence from target domain & (T, W, H)\\
$\boldsymbol{x}_t \in \mathcal{X}$ & Frame from a video in the source domain & (W, H)\\
$\boldsymbol{y}_t \in \mathcal{Y}$ & Frame from a video in the target domain & (W, H)\\
$\bar{z} \in \bar{\mathcal{Z}}$ & Style image from target domain & (W, H)\\
$s$ & State, sequence of frames & (T, W, H) \\
$a$ & Action, next frame of state $s$ & (W, H) \\
$r$ & Reward for the action taken & (1) \\
$s'$ & Next state after taking action $a$ & (T+1, W, H) \\
$\mu$ & Policy networks $ \mu = \{G_x, G_y, P_x, P_y\}$ & (T, W, H) $\rightarrow$ (W, H)\\
$G_x$ & Generator from target to source domain: $\boldsymbol{y} \rightarrow \boldsymbol{x}$ & (T, W, H) $\rightarrow$ (T, W, H) \\
$G_y$ & Generator from source to target domain: $\boldsymbol{x} \rightarrow \boldsymbol{y}$ & (T, W, H) $\rightarrow$ (T, W, H) \\
$P_x$ & Predictor in source domain: $\boldsymbol{x}_{:t} \rightarrow \boldsymbol{x}_{t+1}$ & (T, W, H) $\rightarrow$ (W, H) \\
$P_y$ & Predictor in target domain: $\boldsymbol{y}_{:t} \rightarrow \boldsymbol{y}_{t+1}$ & (T, W, H) $\rightarrow$ (W, H) \\
$Q(s,a) $ & Q-network & (T, W, H ) $\rightarrow$ (1)\\
$D(s,a) $ & Discriminator & (T, W, H) $\rightarrow$ (1)\\
$D_x$ & Discriminator for source domain frames: $x \rightarrow [0,1]$ & (W, H) $\rightarrow$ (1) \\
$D_{\boldsymbol{x}}$ & Discriminator for source domain videos: $\boldsymbol{x} \rightarrow [0,1]$ & (T, W, H) $\rightarrow$ (1) \\
$D_y$ & Discriminator for target domain frames: $y \rightarrow [0,1]^2$ & (W, H) $\rightarrow$ (2) \\
$D_{\boldsymbol{y}}$ & Discriminator for target domain videos: $\boldsymbol{y} \rightarrow [0,1]^2$ & (T, W, H) $\rightarrow$ (2) \\
\bottomrule
\end{tabular}
}
\end{sc}
\end{small}
\end{center}
\end{table}
\end{document}